\documentclass[11pt, a4paper, logo, copyright, nonumbering]{mlins_template}
\usepackage[numbers, sort&compress, round, comma]{natbib}
\usepackage{dblfloatfix}
\usepackage{ulem}
\usepackage{ccaption}
\usepackage{dramatist}
\usepackage{xspace}
\usepackage{pifont} %
\usepackage{multirow}
\usepackage{tcolorbox}
\usepackage{xltabular}
\usepackage{longtable}
\usepackage{hyperref}
\usepackage{bbm}
\usepackage{amsfonts}
\usepackage{amsmath}
\usepackage{amssymb}
\usepackage{lineno}
\usepackage{multirow}
\usepackage{adjustbox}
\usepackage{soul}
\usepackage[bottom]{footmisc}

\usepackage{CJKutf8}
\usepackage{subfigure}
\usepackage{setspace}

\usepackage{dsfont}
\usepackage{array} %
\usepackage{tabularx} %
\usepackage{subfigure} %
\usepackage{xcolor} %
\usepackage{tabularx}
\usepackage{booktabs}
\usepackage{algorithm}
\usepackage{listings}
\usepackage{parskip}
\usepackage{lipsum}  %
\usepackage{multicol} %
\usepackage{lineno}

\makeatletter
\def\@BTrule[#1]{%
  \ifx\longtable\undefined
    \let\@BTswitch\@BTnormal
  \else\ifx\hline\LT@hline
    \nobreak
    \let\@BTswitch\@BLTrule
  \else
     \let\@BTswitch\@BTnormal
  \fi\fi
  \global\@thisrulewidth=#1\relax
  \ifnum\@thisruleclass=\tw@\vskip\@aboverulesep\else
  \ifnum\@lastruleclass=\z@\vskip\@aboverulesep\else
  \ifnum\@lastruleclass=\@ne\vskip\doublerulesep\fi\fi\fi
  \@BTswitch}
\makeatother

\addto\extrasenglish{
}

\newcommand{\tsc}[1]{\textsuperscript{#1}}

\renewcommand{\emph}[1]{\textit{#1}}

\renewcommand{\phi}{\varphi}

\renewcommand{\epsilon}{\varepsilon}
\renewcommand{\imath}{\mathrm{i}}

\newlength{\restsubwidth}
\newlength{\restsubheight}
\newlength{\restsubmoreheight}
\newcommand{\rest}[2]{%
        \settowidth{\restsubwidth}{\ensuremath{#2}}
        \settoheight{\restsubheight}{\ensuremath{{}_{#2}}}
        \ensuremath{{#1\hskip 0.5pt}_{\vrule\kern2pt\parbox[b][%
        4pt][b]{\the\restsubwidth}{%
                        \ensuremath{{}_{#2}}}}}
        }

\renewcommand{\today}{}

\title{\centering Learning neuroimaging models from health system-scale data}

\author{
\centering
Yiwei Lyu\tsc{1}\tsc{*}\quad
Samir Harake\tsc{2}\tsc{*}\quad
Asadur Chowdury\tsc{2}\tsc{*}\quad
Soumyanil Banerjee\tsc{2}\quad
Rachel Gologorsky\tsc{2}\quad
Shixuan Liu\tsc{1}\quad
Anna-Katharina Meissner\tsc{3}\quad
Akshay Rao\tsc{2}\quad
Chenhui Zhao\tsc{1}\quad
Akhil Kondepudi\tsc{2,6}\quad
Cheng Jiang\tsc{2,6}\quad
Xinhai Hou\tsc{2,6}\quad
Rushikesh S. Joshi\tsc{2}\quad
Volker Neuschmelting\tsc{3}\quad
Ashok Srinivasan\tsc{4}\quad
Dawn Kleindorfer\tsc{5}\quad
Brian Athey\tsc{6}\quad
Vikas Gulani\tsc{4}\quad
Aditya Pandey\tsc{2}\quad
Honglak Lee\tsc{1}\quad
Todd Hollon\tsc{1, 2, 6}
\\
\small
\tsc{1}University of Michigan Computer Science and Engineering\quad 
\tsc{2}University of Michigan Neurosugery\quad
\tsc{3}University of Cologne Neurosugery\quad
\small
\tsc{4}University of Michigan Radiology\quad
\tsc{5}University of Michigan Neurology\quad
\tsc{6}University of Michigan Computational Medicine and Bioinformatics\quad \\
\small
\tsc{*}Authors contributed equally \\
\href{https://prima.mlins.org/}{prima.mlins.org}}

\begin{abstract}
Neuroimaging is a ubiquitous tool for evaluating patients with neurological diseases. The global demand for magnetic resonance imaging (MRI) studies has risen steadily, placing significant strain on health systems, prolonging turnaround times, and intensifying physician burnout \cite{Chen2017-bt, Rula2024-qp-1}. These challenges disproportionately impact patients in low-resource and rural settings. Here, we utilized a large academic health system as a data engine to develop \textit{\textbf{Prima}}, the first vision language model (VLM) serving as an AI foundation for neuroimaging that supports real-world, clinical MRI studies as input. Trained on over 220,000 MRI studies, Prima uses a hierarchical vision architecture that provides general and transferable MRI features. Prima was tested in a 1-year health system-wide study that included 30K MRI studies. Across 52 radiologic diagnoses from the major neurologic disorders, including neoplastic, inflammatory, infectious, and developmental lesions, Prima achieved a mean diagnostic area under the ROC curve of $92.0 \pm 5.5\%$, outperforming other state-of-the-art general and medical AI models \cite{Radford2021-wf, Moor2023-oa, Li2023-ru}. Prima offers explainable differential diagnoses, worklist priority for radiologists, and clinical referral recommendations across diverse patient demographics and MRI systems. Prima demonstrates algorithmic fairness across sensitive groups and can help mitigate health system biases, such as prolonged turnaround times for low-resource populations. These findings highlight the transformative potential of health system-scale VLMs and Prima's role in advancing AI-driven healthcare.
\end{abstract}

\setcitestyle{square}

\begin{document}
\maketitle


\clearpage


\section*{Main}
Health systems function as powerful data engines for developing medical foundation models \cite{Moor2023-av, Jiang2023-zr}. Routine clinical operations generate vast volumes of electronic medical records, which can be used to train medical VLMs in a manner analogous to the way internet-scale data is used to train VLMs such as CLIP \cite{Radford2021-wf}, DALL-E \cite{Ramesh2021-cv}, and Flamingo \cite{Alayrac2022-jg}. Globally, approximately 100 million MRI studies are performed annually, with 20\% to 30\% focused on neurological diseases. The demand for brain MRI studies surpasses the available neuroradiology services \cite{Dreisbach2001-eq, Rula2024-qp-1, Christensen2022-tg}. 
This imbalance has caused significant healthcare challenges, including workforce shortages, increased workloads, burnout, and more diagnostic errors \cite{Fawzy2023-it, Krupinski2010-su, Ivanovic2022-sk, Ivanovic2023-nm, ONeill2021-rt, Shin2023-lg, Alexander2022-ly}. Additionally, health disparities in radiology have been exacerbated due to limited resources and a contracting workforce \cite{DeBenedectis2022-oq}. Innovative technologies are needed to improve patient access to radiology services, especially in rural areas and low/middle-income countries. A synergistic collaboration between AI and health systems is essential to address these challenges and improve healthcare delivery.

\begin{figure*}[t]
    \centering
    \includegraphics[width=\textwidth]{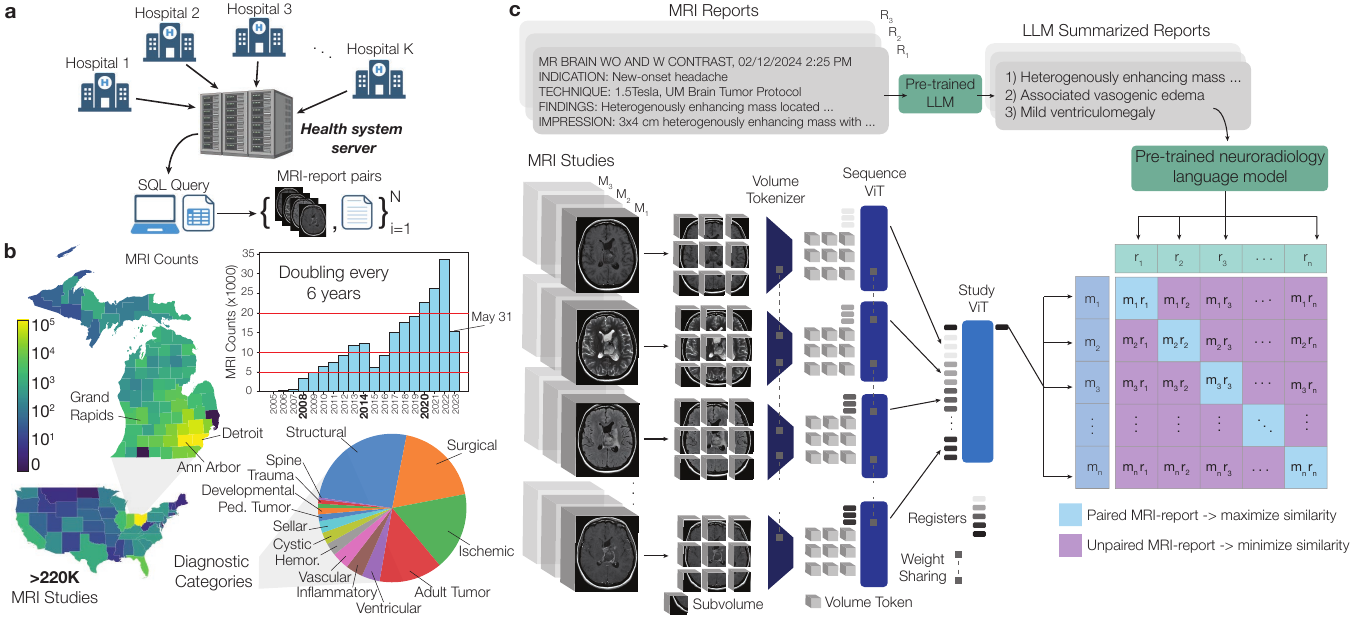} 
    \vspace{-15pt}
    \caption{\small \textbf{Overview of the UM220K MRI Dataset and Prima Workflow.} 
    \textbf{a}, Over 220,000 brain MRIs were queried from our health system's picture archiving and communication system (PACS), forming the UM220K dataset. This dataset includes MRI studies from multiple medical centers across the state and the United States. 
    \textbf{b}, The distribution of MRI counts by county and state is presented. The number of MRIs archived in the PACS system has doubled approximately every six years over the past two decades, highlighting the growing demands on radiology and clinical services. The diagnostic categories reflect the standard operations of a large academic medical center. 
    \textbf{c}, Prima was trained using a contrastive language-image pre-training (CLIP) framework and a hierarchical vision transformer (ViT) architecture. Full MRI studies were divided into subvolumes, compressed into volume tokens using a tokenizer, and processed by a sequence ViT to extract sequence-level features. Global sequence registers were passed to a study ViT to generate a study-level representation for alignment with radiology reports. Radiology reports were summarized using a large language model (LLM), and a pre-trained neuroradiology language model generated report representations. Finally, the MRI study embeddings and summarized report embeddings were aligned using a CLIP objective.}
    \label{fig:workflow}
    \vspace{-15pt}
\end{figure*}

Prima is a general-purpose volumetric MRI VLM trained on health system-scale data, forming a foundation for addressing diverse radiologic and clinical prediction tasks. Traditional approaches to applying AI to MRI studies have relied on manually curated subsets of MRI sequences, such as the FLAIR sequence for lesion detection or T1-weighted images for dementia prediction \cite{Gauriau2021-gr, barbano2024anatomical}. These models are limited by partial radiologic information compared to a radiologist's interpretation of all MRI sequences. Like a radiologist, Prima integrates information from the clinical context, study indication, and all MRI sequences to produce a comprehensive vector representation of the full study, enabling better performance across a broad range of prediction tasks. We demonstrate that Prima's learned representations perform strongly across multiple radiologic, clinical, and biomedical research tasks. This versatility highlights Prima’s potential in optimizing neuroimaging workflows, enhancing diagnostic accuracy, and addressing systemic healthcare challenges.

\begin{figure*}[t!]
    \centering
    \includegraphics[scale=0.8]{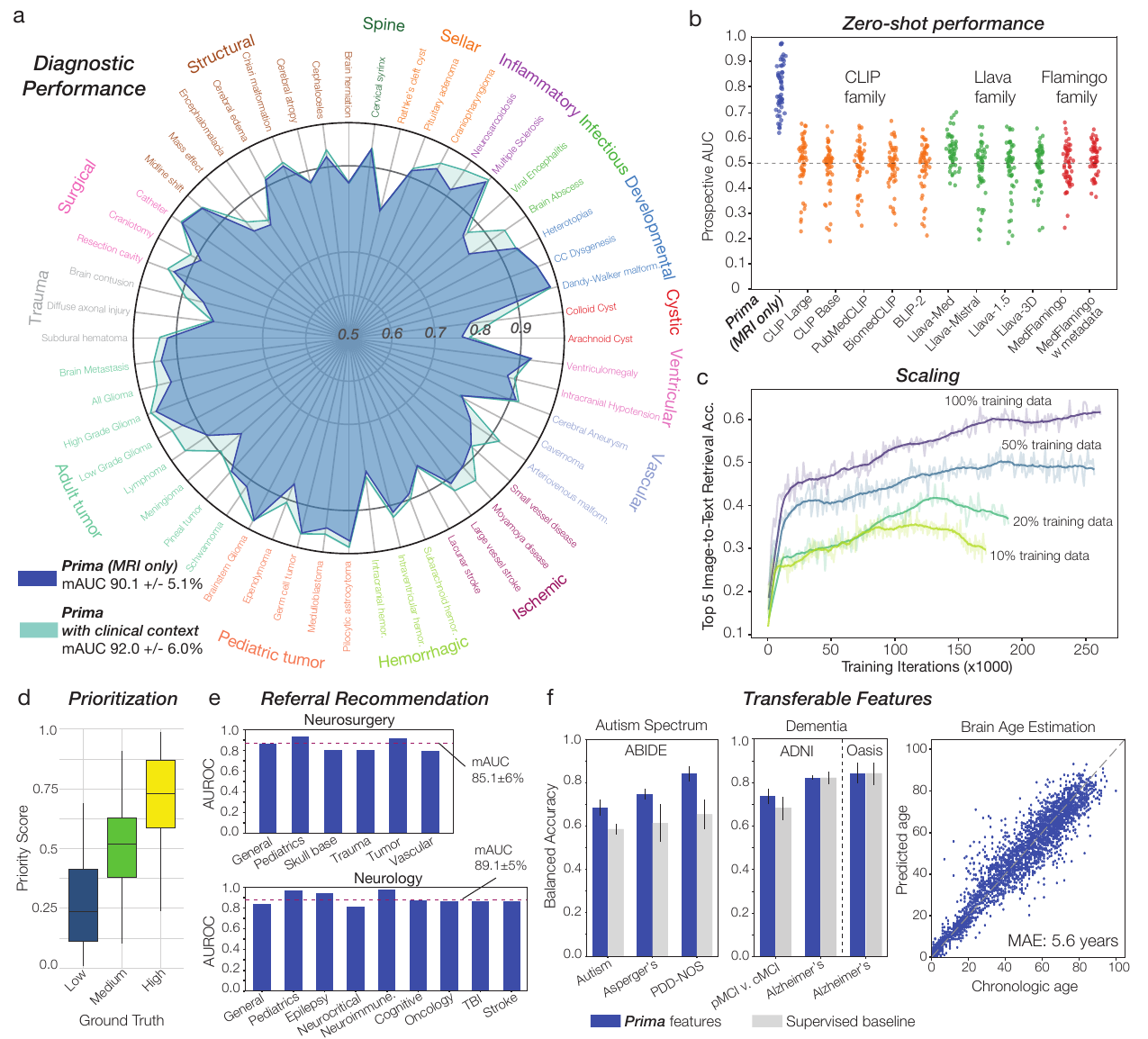}
    \vspace{-15pt}
    \caption{\small \textbf{Overall Performance.} 
    \textbf{a}, Radar plot of AUROC values for each diagnostic task in the testing cohort. Prima demonstrates strong performance across all major diagnostic categories. Similar to radiologists, Prima's diagnostic performance improves given the clinical context/indication.
    \textbf{b}, Vision-language alignment was evaluated via zero-shot classification performance. Each of the 52 diagnoses is shown as a dot for each model. MRI-only Prima outperforms both general and medical VLM models. 
    \textbf{c}, Prima’s generalization performance improves consistently with increased training data, consistent with neural network scaling laws \cite{Kaplan2020-ia}. Performance remains non-saturating with respect to training set size, suggesting additional data will further enhance performance. \textbf{d}, Box-and-whisker plots illustrating the prioritization of MRIs based on diagnostic findings. Prima’s priority scores correlate strongly with ground truth severity scores, yielding a Pearson’s correlation coefficient of $\rho = 0.69$ (95\% CI: 0.68–0.70, $P < 0.001$). \textbf{e}, AUROC and mAUC values for referral recommendations in neurosurgery and neurology specialties, demonstrating high performance. \textbf{f}, Prima’s ability to transfer features to out-of-domain tasks. Bar charts present cross-validation results (mean $\pm$ standard deviation) for autism spectrum disorder and dementia classification, where Prima outperforms \emph{supervised} baselines from the literature \cite{barbano2024anatomical, saratxaga2021mri}. Scatter plot displays brain age estimation results across the lifespan, based on clinical MRIs without exclusions.}
    \label{fig:results}
    \vspace{-15pt}
\end{figure*}

\subsection*{\large Health system-scale vision language models}
To create a large, diverse neuroimaging dataset for VLM development, we queried our health system’s picture archiving and communication system (PACS) for all brain MRIs on May 31, 2023 (Fig. \ref{fig:workflow}, Supplementary Data Fig. 1). After data curation and quality assurance (Extended Data Fig. \ref{exfig:ex_data1}a), the UM220K neuroimaging dataset contained 221,147 MRI studies with paired radiology reports (Extended Data Fig. \ref{exfig:ex_data1}a) from over 170,000 patients. UM220K contains 5.6 million MRI sequences, 362 million MRI slices, and 3.2 billion volume tokens (Fig. \ref{fig:workflow}). UM220K is the largest MRI dataset and includes \emph{all patients} treated or referred to our health system and/or affiliated hospitals since the start of radiology digitization over two decades ago. We aimed to collect a neuroimaging dataset representative of the diverse patient populations and demographics encountered by tertiary health systems delivering primary and specialty care for the full spectrum of neurologic diseases (Extended Data Fig. \ref{exfig:ex_data2}). Manually annotating data at this scale is not feasible. With expert-engineered prompts and radiology reports (Supplementary Data Fig. 2), we leveraged HIPAA-compliant GPT-4 to label MRI studies for 52 radiologic diagnoses from the major neurologic disorders, including neoplastic, inflammatory, infectious, and vascular lesions \cite{OpenAI2023-ny}. Our labeling strategy focused on selecting a diverse, clinically actionable subset of diagnoses to showcase Prima’s ability to learn from health system data. The LLM achieved an average annotation accuracy of 94.0 ± 1.1\%, comparable to expert human annotators across diagnostic categories (Extended Data Fig. \ref{exfig:ex_data3}b).

We designed a hierarchical vision model to align with the MRI data structure, encompassing anatomic regions, MRI sequences, and full studies. Prima’s modular components were trained in three stages: volume tokenization, sequence/study feature learning, and transfer learning for downstream tasks (Extended Data Fig. \ref{exfig:ex_data1}b). First, each MRI sequence was divided into subvolumes (Fig. \ref{fig:workflow}). Inspired by the success of language tokenization and latent diffusion models \cite{Rombach2021-sv}, these subvolumes were transformed into latent volume tokens using a 3-dimensional vector quantized-variational autoencoder (VQ-VAE). The VQ-VAE volume tokenizer is trained at a 16X compression rate using an $\mathcal{L}_1$ reconstruction objective. We achieved high-quality subvolume reconstructions across various MRI sequences, orientations, and pathologies (Extended Data Fig. \ref{exfig:ex_data4}). Compressed volume tokens were pre-saved following model convergence for efficient downstream sequence and study-level training.

The volume tokens from individual MRI sequences were then input into an MRI sequence vision transformer, $ViT_{seq}$ \cite{Dosovitskiy2020-xs}. 3-dimensional position embeddings were concatenated with volume token features to encourage position-aware feature learning. $ViT_{seq}$ is a multimodal vision transformer in two ways. First, the parameters are shared across all sequence types, orientations, and protocols acquired during standard clinical operations. Secondly, $ViT_{seq}$ uses the sequence description metadata, such as `AX\_T2' or `\_plus\_COR\_MPR', as a free text prompt for better feature extraction (Extended Data Figure \ref{exfig:ex_data3}c). Finally, a set of registers were concatenated to the multimodal text-volume token sequence. Vision transformers are known to store global discriminative features in register tokens that can be used for downstream tasks \cite{Darcet2023-sn}. The output registers from each MRI sequence are then input into a study vision transformer, $ViT_{st}$. The $ViT_{st}$ aggregates MRI features using self-attention across the sequence registers. Study classification tokens were used to obtain the full MRI study representation.

Prima is trained using a contrastive language-image pre-training (CLIP) framework \cite{Radford2021-wf}. The objective is to align a full MRI study representation with its corresponding radiology report \cite{Zhang2020-jb, Tiu2022-fn, Bannur2023-qp, Wang2024-mf}. However, raw radiology reports contain textual information that can minimize a CLIP objective but are \emph{not important} for downstream diagnostic tasks, such as protocol information or radiologists' word choices. Moreover, radiology reports can be a source of bias and reduce algorithmic fairness \cite{Chen2023-yy}. An LLM, HIPAA-compliant GPT3.5-turbo, was prompted to summarize each report to distill and itemize the most important diagnostic findings, improving representation learning while minimizing bias and distribution shift (Supplementary Data Fig. 1). $ViT_{seq}$ and $ViT_{st}$ are trained jointly using summarized radiology report supervision with a pre-trained GPT-2 language model as the text CLIP text encoder \cite{radford2019language}. Leveraging the inherent MRI data structure, we added a patient sequence discrimination objective that encourages the $ViT_{seq}$ sequence representations to be similar within a patient's MRI study. MRI features such as brain morphology or pathologic lesions will be shared across MRI sequences and should have similar representations. The patient discrimination objective enforces that these shared features are consistently represented across sequences and improves model convergence (Extended Data Fig. \ref{exfig:ex_data5}e).

\begin{figure*}[t!]
    \centering
    \includegraphics[scale=0.75]{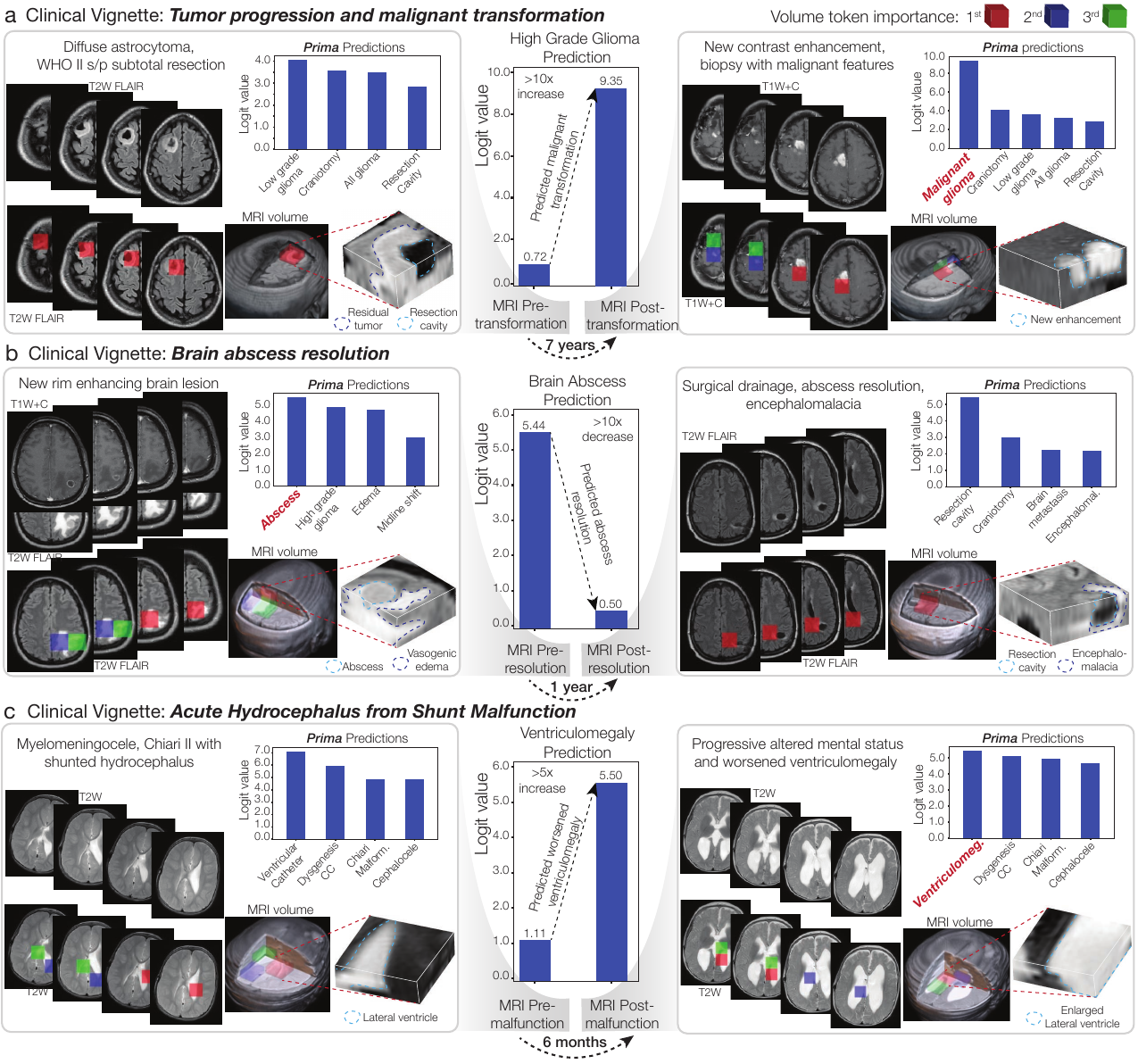} 
    \caption{\small \textbf{Explainable Prima Predictions in Clinical Context.} 
    Three clinical vignettes demonstrate Prima’s explainability using Local Interpretable Model-Agnostic Explanations (LIME). The left panels show patient MRIs at initial presentation (MRI Pre) with the top Prima logits and the top-3 volume tokens identified by LIME. The center bar charts depict changes in Prima logits between the initial presentation (MRI Pre) and after progression or intervention (MRI Post). The right panels display patient MRIs following their clinical courses (MRI Post). 
    \textbf{a}, Clinical vignette of a diffuse low-grade glioma patient, status post (s/p) subtotal resection, who experienced tumor progression and malignant transformation seven years after treatment. Prima accurately identified new regions of contrast enhancement, consistent with malignant glioma. 
    \textbf{b}, Clinical vignette of a patient with a spontaneous brain abscess who underwent surgical drainage and antibiotic treatment, resulting in resolution. 
    \textbf{c}, Clinical vignette of a pediatric patient with a history of myelomeningocele and shunted hydrocephalus. At baseline, the patient had mild ventriculomegaly but presented with acute hydrocephalus following shunt malfunction. Prima accurately predicted the worsening of ventriculomegaly. Interactive demonstration can be found at \href{https://prima.mlins.org/}{prima.mlins.org}.}
    \label{fig:lime}
    \vspace{-15pt}
\end{figure*}

\begin{figure*}[t!]
    \centering
    \vspace{-30pt}
    \includegraphics[width=\textwidth]{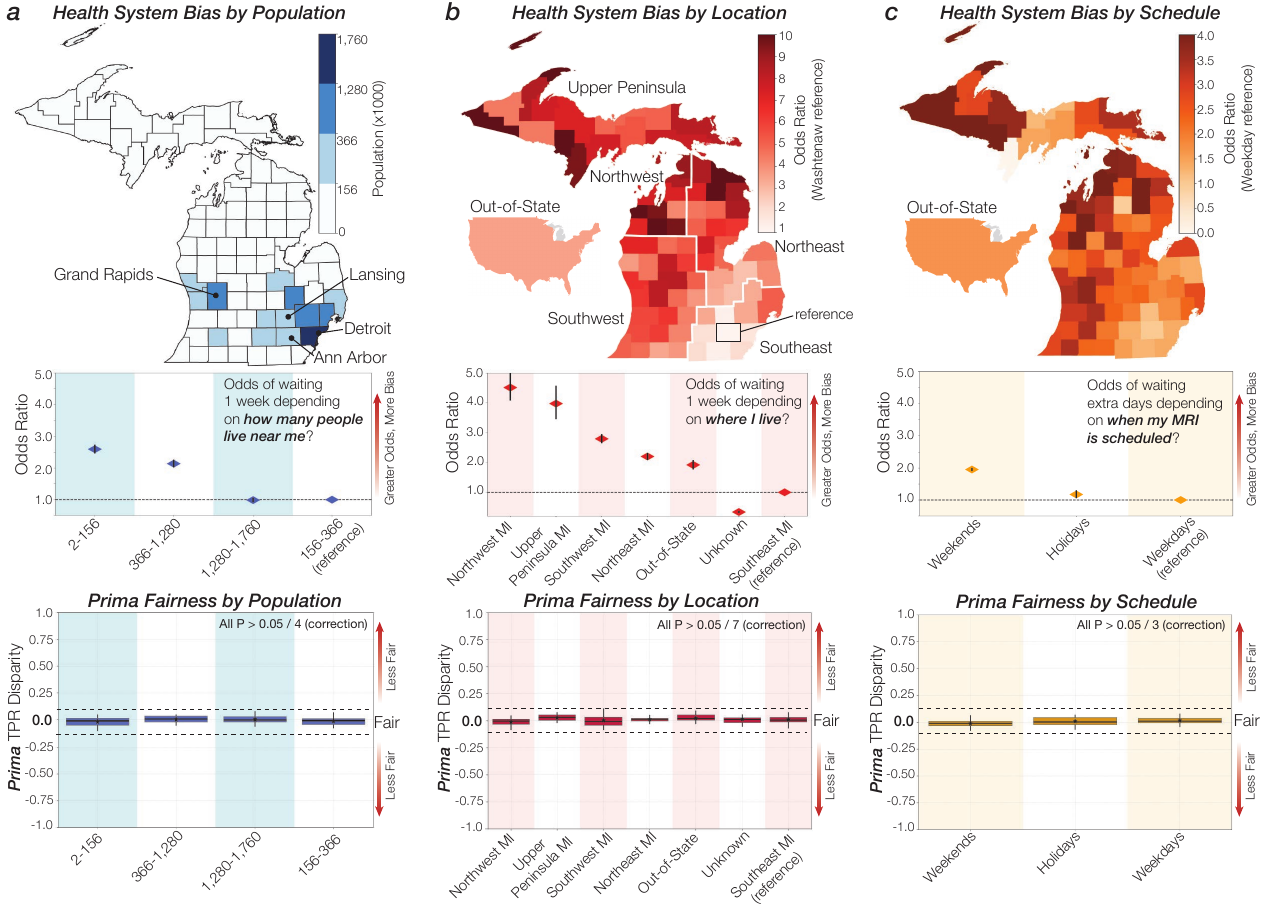} 
    \vspace{-20pt}
    \caption{\small \textbf{Health System Bias and Algorithmic Fairness.} 
    \textbf{a}, State map displaying county populations, grouped into quartiles with equal population sizes. Odds ratios for patients experiencing a 7-day turnaround time are shown for each quartile. Systemic biases were observed, particularly in sparsely populated regions (P $<$ 0.001). Prima demonstrated algorithmic fairness, with minimal true positive rate (TPR) disparity across these population groups. 
    \textbf{b}, State map illustrating counties’ odds ratios for a 7-day turnaround time based on location. The odds ratio plot is further divided by state regions. Systemic biases were prominent in rural areas, especially in Northwest Michigan and the Upper Peninsula (P $<$ 0.001). Despite these biases, Prima maintained consistent fairness across the state and United States. 
    \textbf{c}, State map showing counties’ odds ratios for MRI scheduling delays exceeding 2 days. Turnaround time biases were identified for weekend and holiday MRIs (P $<$ 0.001). Prima exhibited minimal TPR disparity across these subgroups. Diamonds represent odds ratios, and black bars indicate 95\% confidence intervals; intervals not intersecting the dotted line are statistically significant. Box-and-whisker plots display quartile values of TPR disparity, with the bold center line indicating the median, black dots representing the mean, and whiskers extending to data within 1.5 × the interquartile range. Black dashed lines denote 10\% TPR disparity.}
    \label{fig:bias}
    \vspace{-15pt}
\end{figure*}

\subsection*{\large Clinical testing of Prima}
We conducted a 1-year, clinical, offline, health system-wide diagnostic study to test Prima. All patients evaluated at our health system and who received a brain MRI between June 1, 2023 and May 30, 2024 were included as study subjects \emph{without exclusion}. Our study was designed to reliably simulate the clinical setting in which Prima would be deployed for patient care. The testing cohort included 29,435 patients, exceeding the minimum calculated sample size of 22,338. The \emph{primary objective} was to assess Prima’s performance in a multi-label differential diagnosis task spanning 52 neuroradiologic diagnoses. Importantly, radiologists interpret MRI studies using the clinical context and study indications. We aimed to include this clinical context to allow Prima to make more informed diagnostic predictions. The clinical history and study indication from the patient's electronic medical record were embedded using an LLM (OpenAI's text-578 embedding-3-small). The clinical context embedding was concatenated with the MRI study features and used to train a multilayer perceptron (MLP) as a 52-way diagnostic classifier. Clinical context-aware Prima achieved a mean area under the receiver operating characteristic curve (AUROC) of $92.1 \pm 5.5\%$, compared to $90.1 \pm 5.0\%$ with MRI-only Prima. AUROC scores ranged from 78.3\% for arachnoid cysts to 99.7\% for high grade gliomas (Fig. \ref{fig:results}a and Extended Data Fig. \ref{exfig:ex_data5}b). 

We compared Prima with state-of-the-art general VLMs, including OpenAI’s CLIP models \cite{Radford2021-wf} and Microsoft’s LLaVa models \cite{Liu2023-ey}. We also compared Prima to state-of-the-art medical VLMs, including PubMedCLIP \cite{eslami2023pubmedclip}, BioMedCLIP \cite{Zhang2023-rs}, and Med-Flamingo \cite{Moor2023-oa}. Prima showed better vision-language alignment by outperformed all models on zero-shot diagnostic performance (Fig. \ref{fig:results}b, Extended Data Fig \ref{exfig:ex_data5}a). We compared Prima and open-source CLIP models using MLP probing, which includes training an MLP classifier on the UM-220K dataset and tested on the clinical cohort. Prima achieved top performance (Extended Data Fig \ref{exfig:ex_data5}b), indicating the value of the UM-220K dataset and CLIP pre-training on 3D images. Foundation models display performance scaling laws with increased dataset size and compute budget \cite{Kaplan2020-ia}. Prima demonstrated consistent performance improvements with larger training datasets and compute budgets (Fig. \ref{fig:results}c). Consistent improvement in MRI-report alignment was observed on Top-1 and Top-5 retrieval metrics. These results demonstrate that Prima has foundation model properties, and reported performance will continue to improve with additional health system training data and larger compute budgets. Additional subgroup, scaling, vision-language alignment, and confidence calibration analysis can be found in Extended Data Fig. \ref{exfig:ex_data6}.

Next, Prima was tested on two clinical tasks: radiologist's worklist prioritization and clinical referral recommendation. A classifier was trained on frozen Prima features to predict ground truth priority and clinical referrals, which were determined based on the radiologic diagnoses (See Supplementary Data Table 3). For example, patients with evidence of subdural hematomas were assigned high priority whereas patients with arachnoid cysts or unremarkable scans were assigned lower priority (Extended Data Fig. \ref{exfig:ex_data9}a).  Prima’s normalized priority scores were strongly correlated with three-tier ordinal priority scores (normal, medium, high), yielding a correlation coefficient of $\rho$ = 0.69 (95\% CI: 0.68–0.70, P $<$ 0.001) (Fig. \ref{fig:results}d, Extended Data Fig. \ref{exfig:ex_data9}a). Prima was then evaluated on referral recommendations to neurology and neurosurgery specialty care based on MRI features. For example, patients with newly diagnosed multiple sclerosis should be referred to a neuroimmunology specialist. Prima achieved an average neurosurgery referral AUROC of $85.1 \pm 6.0\% $ and neurology referral AUROC of $89.1 \pm 5.0\% $ (Fig. \ref{fig:results}e). These results demonstrate how Prima can improve workflows and streamline clinical care. 

\subsection*{\large Transferable Prima features}
CLIP-learned visual representations can transfer effectively to various downstream tasks, including out-of-distribution scenarios \cite{Radford2021-wf}. We evaluated Prima's MRI representations using a linear evaluation protocol on three benchmarked neuroimaging tasks: autism spectrum prediction~\cite{di2014autism}, dementia/Alzheimer's disease prediction \cite{petersen2010alzheimer, marcus2007open}, and brain age estimation \cite{Lee2022-ec}. These tasks are considered out-of-distribution because the summarized radiology reports lack information about the patient’s clinical diagnoses or age. For autism spectrum and dementia predictions, Prima matched or exceeded performance of independent fully \emph{supervised and semi-supervised} benchmarks on three publicly available datasets: ABIDE~\cite{di2014autism}, ADNI~\cite{petersen2010alzheimer}, and OASIS~\cite{marcus2007open} (Fig. \ref{fig:results}f) \cite{barbano2024anatomical}. Using Prima features, brain age estimation yielded a mean absolute error (MAE) of 5.6 years on our testing dataset. These results are competitive with existing models trained end-to-end for brain age estimation on large, uncurated, clinical cohorts \cite{Bashyam2020-de}. Our findings also demonstrate that Prima's performance transfers effectively to other public datasets and diagnostic tasks, including diffuse gliomas classification (BRATS~\cite{baid2021rsna}), brain metastasis prediction (UCSFMets~\cite{rudie2024university}, NYUMets~\cite{oermann2023longitudinal}), and acute strokes detection~\cite{liu2023large} (Extended Data Fig. \ref{exfig:ex_data9}b). Notably, prior models require extensive preprocessing, including skull stripping, sequence selection, resampling, and segmentation. Prima’s flexible architecture enables predictions from \emph{any sequence or sequence combination without preprocessing}.

\subsection*{\large Explainable Prima predictions}
Explainable AI is essential in healthcare to ensure safe, reliable, and trustworthy predictions \cite{Wiens2019-mh}. We assessed Prima’s predictions using Local Interpretable Model-Agnostic Explanations (LIME) \cite{ribeiro2016should}. LIME assigns importance scores to individual volume tokens in an MRI, with higher scores indicating greater contributions to Prima’s predictions. If LIME highlights pathologic regions in an MRI with high importance scores, then Prima’s diagnostic predictions are aligned with clinical reasoning. Fig. \ref{fig:lime} showcases three clinical vignettes illustrating Prima’s value with LIME explanations: malignant brain tumor transformation, brain abscess resolution, and acute hydrocephalus due to shunt malfunction. Each vignette illustrates Prima’s ability to generate accurate and trustworthy predictions across a patient’s clinical course. For example, LIME revealed that Prima accurately identified regions of new contrast enhancement—a well-established radiologic marker of malignant transformation \cite{Smith2008-ga}—to predict the progression of a low-grade glioma into a malignant glioma. We quantitatively validated Prima by assessing its ability to assign high LIME scores to manually segmented brain tumor regions in the expert-annotated BraTS dataset. Prima achieved 98.0\% Top-3 accuracy in selecting tokens within segmented brain tumor regions (Extended Data Fig. \ref{exfig:ex_data9}d). Extended Data Fig. \ref{exfig:ex_data7} demonstrates LIME visualizations for various pathologies, including pediatric, inflammatory, infectious, and developmental lesions, underscoring Prima’s versatility. As a multi-label classifier, we performed a multi-label analysis to demonstrate that Prima learns the co-occurrence between correlated diagnoses, such as brain contusion and midline shift (Extended Data Fig. \ref{exfig:ex_data8}a). Prima selects different volume tokens when making different diagnostic decisions \emph{for the same patient and MRI study}. Prima correctly selects volume tokens in the posterior fossa when diagnosing a pediatric cerebellar brain tumor and selects tokens in the lateral ventricles to diagnose associated ventriculomegaly from obstructive hydrocephalus (Extended Data Fig. \ref{exfig:ex_data8}b).

\subsection*{\large Health system bias and Prima fairness}
Health system bias and health disparities are pervasive across all medical specialties \cite{DeBenedectis2022-oq, Waite2021-st}. Ensuring algorithmic fairness is critical for medical AI models to achieve equitable performance across sensitive attributes and mitigate existing disparities \cite{Chen2023-yy}. To assess Prima's algorithmic fairness, we examined a key modifiable source of health system bias in radiology—\emph{turnaround time}. Turnaround time is the interval between when an imaging exam is performed and when the radiologist's report is accessible to the referring health care provider. Quick turnaround times are critical as timely diagnosis can impact patient care. Importantly, turnaround time is influenced by various health system factors, including imaging study complexity, radiologist workload, the need for specialized interpretation, and overall health system efficiency. Final interpretation timestamps were used to calculate turnaround times in UM-220K. Due to increasing radiology volumes, \emph{average turnaround time at our health system has increased over time} from a low of approximately 18 hours in 2012 to a high of over 2.25 days in 2024 (Extended Data Fig. \ref{exfig:ex_data2}b). The turnaround time distribution showed a large right skew toward longer turnaround times, with the majority of turnaround time measured in patient-days found in the right tail of the distribution (Extended Data Fig. \ref{exfig:ex_data2}c). We identified three sensitive attributes that account for the tail distribution and lead to systemic biases affecting turnaround time: population density, geographic region, and scheduling. Patients in sparsely populated rural areas were 2 to 5 times more likely to experience a 7-day turnaround time compared to those in urban areas (P $<$ 0.001, Fig. \ref{fig:bias}a). Patients who had an MRI scheduled during a weekend were twice as likely to wait three or more days (P $<$ 0.001, Fig. \ref{fig:bias}a). Because turnaround time is negligible for Prima, $< 3$ seconds on single graphical processing unit, we evaluated Prima’s algorithmic fairness across sensitive groups to mitigate systemic biases. In Fig. \ref{fig:bias}, we compared the true positive rate (TPR) of sensitive groups with the study population, known as equalized opportunity \cite{Barocas2023-nn}. Larger TPR disparity between the sensitive group and the study population indicates worse algorithmic fairness. Prima exhibited algorithmic fairness across the three primary sources of health system bias (Fig. \ref{fig:bias}). Further subgroup performance, equalized odds, and intersectional analysis revealed equitable performance across patient risk factors (e.g., race, sex, age), medical access factors (e.g., insurance status, MRI manufacturer), reinforcing Prima’s robustness and fairness (Extended Data Fig. \ref{exfig:ex_data10} and Supplementary Data Fig. 6).









\section*{Discussion}
Prima is a general-purpose neuroimaging VLM trained on health system-scale data, delivering general, scalable, and equitable performance. Leveraging over 220,000 MRI studies (over 5.6 million 3D sequences) from diverse patient populations, Prima establishes a new benchmark in radiologic diagnosis and clinical prediction. Our study underscores the potential of health system-scale models to improve clinical efficiencies and ease labor shortages. Unlike earlier neuroimaging models that rely on curated datasets and pre-selected sequences, Prima excels with large, uncurated imaging data, making it highly practical for real-world AI applications.

A phased certification pathway has been proposed for evaluated generalist medical AI systems that mirror the clinical training of physicians \cite{Rajpurkar2025-mm}. The first stage is evaluating AI systems for baseline competency through standardized testing and scenario analysis. Our study is limited to addressing only this initial stage of Prima’s clinical certification. Reported diagnostic accuracy for expert neuroradiologists is over 94\% \cite{Ivanovic2023-ny, Babiarz2012-ui, Wu2014-wl}. By evaluating via standardized testing benchmarks and clinical tasks, we aimed to show only that Prima has established a foundation of medical knowledge. Future work will explore integrating detailed clinical notes and EHR data as input, advanced VLM tasks, including open-ended diagnosis, automated report generation, and visual question answering. We aim for Prima's series and study representations to be seamlessly aligned with large language models, enabling radiologist-level performance on complex interpretation tasks.

The broader impact of Prima extends beyond neuroimaging. Our proposed AI framework is broadly adaptable to other biomedical imaging modalities, such as computed tomography, radiography, and ultrasound. We hope our proposed framework can contribute to existing medical AI models for other organ systems and imaging modalities \cite{Azizi2023-xp, moor2023medflamingo, Singhal2023-rt, Blankemeier2024-dw}. Prima's ability to provide comprehensive representations of clinical MRIs holds promise for advancing research in neuroimaging. Immediate research applications of Prima include brain phenotyping \cite{Elliott2018-vz} and quantifying disease progression and treatment response \cite{Kickingereder2019-ey}. Future model versions will incorporate genetic and clinicopathologic patient data to further improve predictions and explore pathophysiological insights.

In conclusion, Prima exemplifies the transformative potential of integrating health systems and medical foundation models to improve healthcare. As healthcare datasets grow and compute resources expand, model performance and utility are poised to scale, offering a pathway to AI-driven innovation in medicine.

\clearpage
\section*{Methods}
\subsection*{Overall objectives and study design} 
The primary objective was to develop, optimize, and evaluate a VLM trained on health system-scale data to achieve general and transferable representations of brain MRI studies. The University of Michigan Institutional Review Board (IRB) approved the collection and analysis of retrospective and prospective MRIs conducted during routine clinical operations at Michigan Medicine. The study included secondary data analysis only whereby existing electronic medical record data was used for model training and testing. We adhered to five study design principles for dataset selection, model architecture, and prediction tasks: 1) inclusive data criteria/`data in the wild', 2) minimize data preprocessing, 3) flexible vision language modeling, 4) multi-modal model input (image and text metadata), and 5) clinically informative and diverse prediction tasks. Due to the heterogeneity of clinical MRI protocols, previous studies have relied on data preprocessing and curation, such as volume resampling or limiting MRI sequence inclusion (e.g. T1 only). This allows for more standardized model inputs, but limits dataset size and downstream prediction tasks. We aimed to be data-inclusive and develop a general vision language modeling strategy to accommodate the full range of clinical MRI study protocols, such as brain, pituitary, epilepsy, perfusion, etc. Moreover, clinical MRI studies have an inherently hierarchical data structure: voxels $>$ regions $>$  sequences $>$ studies. We have previously demonstrated that leveraging hierarchical data structures can improve representation learning in biomedical imaging \cite{Jiang2023-rq}. We developed hierarchical vision transformers for brain MRIs studies to achieve high-quality and transferable representations using radiology report supervision \cite{Radford2021-wf}. We performed a health system-scale clinical study of Prima performance for rigorous and reproducible testing. The major aim of the real-world, clinical testing was to demonstrate that Prima can provide preliminary radiologic diagnoses, study triage/worklist prioritization, and referral recommendations from MRIs studies alone. Full overview of the training and inference workflow can be found in Extended Data Figure \ref{exfig:ex_data1}. Finally, we aimed to show the algorithmic fairness of Prima and present those results in the context of known health system-level biases that results in healthcare disparities.

\subsection*{Data curation of UM-220K MRI dataset} 
Large-scale MRI data acquisition and curation was essential for study feasibility \cite{Wood2024-sl, Ghosh2024-tz}. Michigan Medicine uses Sectra PACS as a third-party vendor for all radiology study viewing, storage, and archiving. Michigan Medicine maintains a SQL-based research interface with our clinical PACS system, called Sectra Data Warehouse (SDW). SQL queries were tailored to identify all MRIs completed through May 30, 2023 that included one or more of the following body parts: head, brain, orbits, face, or neck. Details of the SQL query can be found in Supplementary Data Figure 1. The query resulted in a total of 279,908 number of hits. We then filtered the MRI dataset to ensure that all query images had (1) associated radiology reports, (2) a minimum of two MRI sequences, and (3) non-corrupted data (Extended Data Figure \ref{exfig:ex_data2}). All studies were pushed to a HIPAA-compliant server mounted on the University of Michigan Armis2 high-performance computing cluster. The MRI sequences were then converted to LPS orientation and images were rescaled to $256 \times 256$ pixels in the X and Y planes, and slice thickness was converted to 4mm or greater in the Z plane.

\subsection*{MRI volume tokenization}
A common processing step when using ViT architectures is splitting full images into smaller image patches, or vision tokens \cite{Dosovitskiy2020-xs}. Most ViTs use image tokens that are flattened 2D image patches of size $16 \times 16$ or $8 \times 8$, for example. Unfortunately, applying this patching strategy to 3D MRI volumes results in a prohibitively large number of tokens per sequence, up to $>$32K for $8 \times 8 \times$ 8 patches. We designed a volumetric tokenization strategy that reduces the number of tokens per MRI sequence while preserving diagnostic features. Inspired by text tokenization and latent diffusion models \cite{Rombach2021-sv}, our strategy trains vision language models in the latent space of a pretrained autoencoder. This strategy, called \textit{volume tokenization}, uses a variational autoencoder objective with vector quantization (VQ-VAE) and 3D convolutional neural networks on MRI volume patches to generate tokens for vision transformer input \cite{Van_den_Oord2017-lk}. Volume tokenization is a compression module that eliminates uninformative details from the 3D regions. This strategy enables efficient vision language model training with limited computational resources while retaining quality and flexibility.

MRI sequences were divided into $32 \times 32 \times 4$ volume patches, $x$, corresponding to the $X \times Y \times Z$ image dimensions, with zero-padding applied as needed. The patched dataset was used to train a VQ-VAE comprising a 3D CNN encoder ($f$), a quantization layer with a codebook of size 8192, and a 3D CNN decoder. The encoder downsampled each patch to an $8 \times 8 \times 2$ volume with 2 feature dimensions, resulting in a compact embedding vector $\mathbf{z}_e \in \mathbb{R}^{256}$ (i.e., $8 \times 8 \times 2 \times 2$) that serves as the input to the vision model. The codebook size was chosen to balance reconstruction quality and computational efficiency (Extended Data Figure \ref{exfig:ex_data4}).

The VQ-VAE model maps the encoder output $ f(x) = \mathbf{z}_e(x)$ to the nearest embedding vector $\mathbf{e}_k$ in the codebook:
\begin{equation}
\mathbf{z}_q(x) = \mathbf{e}_k \quad \text{where } k = \arg\min_j \lVert \mathbf{z}_e(x) - \mathbf{e}_j \rVert_2
\end{equation}
We favor using vector quantization via a discrete codebook because anatomic structures and pathologic features are often shared across patients and pathologies. We demonstrate in Extended Data Figure \ref{exfig:ex_data4} that normal structures and radiographic diagnoses have similar embeddings both within and across patients. To ensure robustness across different imaging planes (e.g., axial, coronal, sagittal) and spatial orientations (e.g. LPS, RAS, LIP, etc), we implemented an MRI sequence transformation that applies a random permutation of the image axes during training. This ensures that the 3D-CNN encoder remains invariant to the imaging plane and orientation. Enforcing this invariance is essential given that clinical MRIs in the UM-220K dataset include multiple planes and orientations. High-quality reconstructions across various imaging planes, spatial orientations, and permutations are shown in Extended Data Figure \ref{exfig:ex_data4}. Volume tokenization effectively mitigates the challenge of handling large token counts in 3D MRI sequences while maintaining essential diagnostic information, enabling scalable and flexible vision language model training.

\subsection*{MRI report summarization}
High-quality natural language annotations improve CLIP training  \cite{Ramesh2022-bv}. For MRI-report pairs, the encoded radiology report serves as the `target' for the vision model during CLIP training. Therefore, the quality of the report target determines the quality of learned MRI features. Uncurated, clinical radiology reports can contain extraneous information that may not contribute to visual representation learning. Moreover, non-diagnostic patterns in radiology reports, such as word choice, grammar, and unique comments, can be a source of data leakage and bias. To improve radiology report targets and minimize data bias, we used GPT-3.5 for radiology report summarization of UM-220K in preparation for CLIP training \cite{Brown2020-dv}. GPT-3.5 is known to provide high-quality, clinical-grade report summarizations for medical imaging \cite{Chien2024-ls, Ranjit2023-hk}. GPT-3.5 was prompted to provide an enumerated list of the most salient diagnostic information, remove extraneous information/details, remove references to normal/unremarkable structures, and remove comparison statements (e.g. stable, progression, previous, improved, etc.). The prompt used for report summarization can be found in Supplementary Data Figure 1. Report summarization aimed to (1) remove report text that may improve retrieval performance without improving classification performance, (2) homogenize reports, and (3) minimize data leakage, bias, or learning spurious correlations between the MRI-report pairs. MRI report summarization leads to better text representations (Extended Data Figure~\ref{exfig:ex_data3}f) and significantly improved downstream classification performance with results shown in Extended Data Figure \ref{exfig:ex_data5}d. In Extended Data Figure~\ref{exfig:ex_data3}d, we show that Prima, though trained on GPT-3.5 summarization, is aligned with summarization from neuroradiology experts.

\subsection*{MRI labeling with LLMs}
We aimed to assign diagnostic labels to each MRI study in UM-220K using the clinical radiology reports. Language models have been used extensively to automate data annotation and radiology report labeling \cite{Adams2023-ob, Titano2018-mx, Chien2024-ls, Ranjit2023-hk}. We selected 52 labels that spanned the full neurological disease spectrum to ensure diverse and clinically important predictive tasks. For each diagnosis, MRI reports were first filtered using keywords and string pattern matching. For example, we would used keywords 'subdural' AND 'hemat' string matching to filter reports for possible subdural hematoma diagnosis. After filtering, we used a standardized prompt to HIPAA-aligned GPT-4 that included the keywords and diagnosis to be labeled (Supplementary Data Fig. 3). GPT-4 was prompted to respond with a binary 'yes' or 'no' for each label. To assess the quality of the automated, `silver standard' annotations, we compared them to `gold standard' annotations from an expert in neuroradiology (A.K.M). We selected a diverse set of diagnostic classes, including meningioma, brain abscess, acute ischemic stroke, Chiari malformations, and arachnoid cyst, to evaluate the quality of GPT-4 annotations. GPT-4 annotation performance results are shown in Extended Data Fig. \ref{exfig:ex_data3}b.

\subsection*{Hierarchical multimodal transformers}
Following MRI volume tokenization, a vision model was trained to learn representations of MRI sequences and studies. We used a two-level hierarchical ViT ($hViT$). The sequence ViT ($ViT_{seq}$) was used to encode MRI sequence features, and the study ViT ($ViT_{st}$) was used to aggregate the sequence features and produce representations of full MRI studies. Capturing the hierarchical data structure of MRIs reduces the series length for both the sequence and study ViT by minimizing the number of input tokens for any ViT forward pass. As shown in Fig. \ref{fig:workflow}, the sequence transformer weights are shared across all the sequences to reduce the model size and improve training efficiency. 

We define each MRI study as $M=(stn,\{s_1,s_2,...,s_m\})$ where $stn$ is the study name, such as `MRI BRAIN WITH AND WITHOUT CONTRAST' and each $s_i$ is a tokenized sequence in the study. Each tokenized sequence $s_i=(sn_i,Z_i)$ where $sn_i$ is the sequence name, and $Z_i=\{z^1_i,z^2_i,...,z^{n_i}_i\}$ is the set of VQVAE-encoded tokens in the sequence, each concatenated with a 30-dimensional sinusoidal positional embedding \cite{Vaswani2017-in} based on its 3D coordinates within the sequence and a 3-dimensional one-hot vector indicating the sequence's original orientation (axial/sagittal/coronal). To improve efficiency and reduce memory requirement, we used pixel intensity filtering to remove the background tokens, and we define the filtering process as $F$, such that $F(Z_i) \subset Z_i$ is the post-filtering token subset. See Extended Data Figure \ref{exfig:ex_data1} for a schematic summary.

Similar to how radiologists interpret voxel intensities differently depending on the MRI sequence, we included the tokenized sequence name as part of the input to the sequence ViT. Sequence names (such as "Ax\_T2\_FLAIR") contain essential information for interpreting voxel intensities for different tissues and pathologies.  We use a sequence name encoder $E_{sn}$, a 3-layer character-level transformer, to encode sequence names. We pre-trained $E_{sn}$ with CLIP objective between $E_{sn}(SN_i)$ and $V(F(Z_i))$, where $V$ is ViT model. Visualizations of the learned series name embeddings can be found in Extended Data Fig. \ref{exfig:ex_data3}c. 

The input sequence to $ViT_{seq}$ for each sequence $s_i$ contains 3 parts: 20 register tokens (trainable parameters), $E_{sn}(SN_i)$, and $F(Z_i)$. The 3 parts are concatenated and input into $ViT_{seq}$. The encoded vector for each sequence ($r_i=ViT_{seq}(s_i)$) is an 1024-dimentional vector obtained by the final layer output over the 20 registers concatenated together along feature dimensions then projected through a linear layer.

The input sequence to $ViT_{st}$ for study $M$ also contains 3 parts: 10 register tokens, $E_{stn}(STN)$, and $\{P(r_1), P(r_2), ... , P(r_m)\}$, where $E_{stn}$ is the study name encoder (same architecture as $E_{sn}$ but not pre-trained) and $P$ is a linear projection layer. The 3 parts are concatenated together, and fed into $ViT_{stu}$. The encoded vector for the entire study M is the final layer output over the 10 register tokens concatenated along feature dimension to form a single 10240-dimensional feature vector (1024 dimensions x 10 registers).

$ViT_{seq}$ has 15 transformer layers, each layer with 16 heads (64 dim each); $ViT_{st}$ has 4 transformer layers, each with 8 heads (64 dim each). The entire visual encoder ($ViT_{seq}$ and $ViT_{st}$ combined, not including VQ-VAE) has 56.584 million parameters.

\subsection*{Training objective on MRI-report pairs}

We train $hViT$ via a CLIP objective~\cite{Radford2021-wf} between the $hViT$ representation and the representations of the corresponding summarized text reports. The text encoding model, $G$, is a GPT-2 model pre-trained on radiology reports using an autoregressive next-word prediction objective. We found that pre-training the text encoder on the radiology report corpus improved VLM training efficiency (Extended Data Fig. \ref{exfig:ex_data5}f). For a batch of $k$ MRI study-report pairs, $B=\{(M_1,R_1),(M_2,R_2),...,(M_k,R_k)\}$, where $R_1,R_2,...R_k$ are summarized reports, the CLIP objective is as follows:

\begin{align}
    v^M_i &= P_M(hViT(M_i)),i \in [1,...,k]\\
    v^R_i &= P_R(G(R_i)) \in [1,...,k]\\
    L_{CLIP_M} &=  \frac{1}{k}\sum_{i=1}^k -\log\frac{\exp(\text{sim}(v^M_i,v^R_i)*exp(\tau))}{\sum\limits_{j=1}^k \exp(\text{sim}(v^M_i,v^R_j)*exp(\tau))}\\
    L_{CLIP_R} &= \frac{1}{k}\sum_{i=1}^k -\log\frac{\exp(\text{sim}(v^M_i,v^R_i)*exp(\tau))}{\sum\limits_{j=1}^k \exp(\text{sim}(v^M_j,v^R_i)*exp(\tau))}\\
    L_{CLIP} &= L_{CLIP_M} + L_{CLIP_R}
\end{align}

Where $sim$ is the cosine similarity, $sim(x,y) = \frac{\mathbf{x} \cdot \mathbf{y}}{\|\mathbf{x}\| \|\mathbf{y}\|}$, and $P_R$ and $P_T$ are linear projection layers with an output dimension of 128. The temperature parameter, $\tau$, was initialized to 0.07 and updated during training. We found that this significantly improved optimization and overall performance. We employed various augmentations on both the reports and MRI images. These included shuffling the order of entries in summarized reports, dropping visual tokens, replacing sequence names with "unk", changing the filtering threshold of $F$, and dropping full sequences. In our UM-220K dataset, approximately 40\% of training MRI studies contain no significant abnormalities. To address dataset imbalance, we upsample the abnormal studies by 4 times during CLIP training to improve efficiency by reducing the number of studies within each CLIP batch that has indistinguishable normal/unremarkable reports.

\subsection*{Self-supervised patient discrimination objective}

In addition to the CLIP objective, we added a self-supervised patient discrimination objective that leverages the hierarchical structure of MRI studies. All sequences from an MRI study are of the same patient; therefore, neuroanatomic features are shared across MRI sequences from that patient. We developed a patient discrimination objective that will enforce that sequences of the same patient will have similar representations from $ViT_{seq}$. Let $s_i^j$ denote the $ViT_{seq}$ representation for the jth sequence in $M_i$ and let $n_i$ denote the number of sequences in $M_i$. The patient sequence discrimination objective is as follows:

\begin{align}
    u^{j}_i &= P_{patdis}(s_i^j)\\
    L_{patdis} &= \frac{1}{k} \sum\limits_{i=1}^k \frac{1}{n_i}\sum_{j=1}^{n_i} -\log\frac{\sum\limits_{j'=1}^{n_i} \exp(\text{sim}(u^{j}_i,u^{j'}_i)/\tau_p)}{\sum\limits_{i'=1}^{k}\sum\limits_{j'=1}^{n_{i'}} \exp(\text{sim}(u^{j}_i,u^{j'}_{i'})/\tau_p)}\label{eq:patdis}
\end{align}

where $P_{patdis}$ is a 2-layer MLP projection layer that maps $ViT_{seq}$ outputs to the patient discrimination embedding space, and $\tau_p$ is a trainable temperature parameter initialized at 0.1. The numerator of equation~\ref{eq:patdis} are the sequence similarities within the same study/patient. The denominator is the sequence similarities between all pairs of sequence representations. The final training objective for Prima is 

\begin{equation}
    L_{train} = L_{CLIP}+\lambda L_{patdis}
\end{equation}

where $\lambda$ is a hyperparameter. In our experiments, we set $\lambda$ to 0.03.

\subsection*{Evaluation metrics for vision language alignment}

During CLIP training, we used Top-1 and Top-5 retrieval accuracies to monitor vision language alignment. These metrics evaluate the accuracy of Prima for matching MRI studies with their corresponding summarized radiology report. Top-1 and Top-5 retrieval accuracy metrics were defined as:

\begin{align}
\text{Top-1 Accuracy} &= \frac{1}{N} \sum_{i=1}^{N} \mathbbm{1}\left( \arg\max_{j} \text{sim}(i, j) = i \right)\\
\text{Top-5 Accuracy} &= \frac{1}{N} \sum_{i=1}^{N} \mathbbm{1}\left( i \in \text{Top-5}\left( \text{sim}(i, j) \right) \right)
\end{align}

where N=254 and is a held-out validation set from UM220K. To evaluate vision language alignment on the clinical testing set, we randomly divide the testing set into 294 groups of 100, and then report top-1 retrieval accuracy within each group, averaged across all groups. 

\subsection*{Evaluation protocol for diagnostic tasks}
Radiologic diagnoses are not mutually exclusive and cannot be treated as a multi-class classification task. Radiologic diagnoses can co-occur with high probability, such as brain tumors and vasogenic edema or brain atrophy and ventriculomegaly. Moreover, while some diagnoses can be made from the imaging study alone, many radiologic diagnoses are \emph{differential diagnoses}, a list of semantically related medical or pathologic conditions that could explain the imaging findings observed in a patient's MRI study. The importance of learning the semantic relationships within differential diagnoses applies to the majority of broad categories in our study, including neoplastic, inflammatory, infectious, and vascular. Therefore, we aimed for the diagnostic model to learn the co-occurrence and semantic relationship of the radiologic diagnoses. We trained a classification head $C$ that takes as input the MRI study embeddings generated by $hViT$, $v^M$. $C$ is a 3-layer multilayer perceptron and outputs a $L$-dimensional vector, where $L$ is the number of labels:
\begin{equation}
    \boldsymbol{\hat{y}} = C(v^M),  \quad
\boldsymbol{\hat{y}} = [y_1, y_2, \dots, y_L], \quad y_i \in [0, 1]
\end{equation}

Similar to other CLIP models \cite{Radford2021-wf}, $hViT$ is fixed and $C$ is trained using a positive-weighted binary cross entropy loss:
\begin{equation}
\mathcal{L}_{\text{BCE}}^{\text{multi-label}} = - \frac{1}{L} \sum_{i=1}^{L} \left[ p_i \, y_i \log(\hat{y}_i) + (1 - y_i) \log(1 - \hat{y}_i) \right]
\end{equation}
with $p_i$ as the positive weight for the $i$th label obtained by $p_i=\frac{\text{num negative ith label in training set}}{\text{num positive ith label in training set}}$. The ground truth multi-hot vector, $y$, is provided by the LLM as described above,
\begin{equation}
\boldsymbol{y} = LLM(R),  \quad
\boldsymbol{y} = [y_1, y_2, \dots, y_C], \quad y_i \in \{0, 1\}
\end{equation}

where $R$ is the MRI study report. We selected the checkpoint $C$ with the best performance for each task using a held-out validation set. We found that the above strategy successfully captures label co-occurence and learns the semantic/differential diagnosis relationships between labels (Extended Data Fig. \ref{exfig:ex_data8}).

\subsection*{Testing patient cohort and sample size calculation}
Our main objective was to test Prima in a real-world, offline, health system-wide, diagnostic accuracy study that included patients in an unbiased and uncurated fashion. We designed a 1-year diagnostic accuracy study and used the same inclusion criteria as shown in Extended Data Fig. \ref{exfig:ex_data1}. We performed a sample size calculation based on a parallel superiority trial with a binary outcome: differentiating normal versus abnormal MRIs. We selected this task based on a recent related study by Gauriau et al. using deep learning for triaging brain MRIs \cite{Gauriau2021-gr}. They acheived an AUROC of 0.78 (95\% confidence interval: 0.75, 0.80). We set the control (Gauriau et al.) and experimental performance (Prima) to 0.78, the superiority limit to 2\% (based on the above CI), alpha to 0.01, and power to 0.90, resulting in a minimal sample size of 22,338 MRIs. The primary evaluation metric was mean AUROC across the radiologic diagnosis tasks. Testing patient cohort enrollment began on June 1, 2023 and completed on May 30, 2024. A total of 29,435 MRI studies were included and our required minimal sample size was exceeded. All sample size calculations were completed in R (4.3.0) using the epiR package (2.0.63) epi.sssupb function.

\subsection*{Model Design Ablations}
We performed several ablation studies to optimize the design choices of the Prima architecture: inclusion of sequence name and study description, flat ViT encoder, 2D CNN encoder, 3D CNN encoder, unsummarized long reports, and self-supervised patient discrimination loss. The results are shown in Extended Figure \ref{exfig:ex_data5}d.

To ablate on the inclusion of sequence name and study descriptions, we take evaluate the performance of Prima model with no sequence name information (i.e. replace all input sequence name with "unk") and no study description (study name encoding not included in the input to $ViT_{st}$). We found a slight drop in performance after removing either components, indicating that the inclusion of sequence name and study description does help with prediction accuracy, but Prima also does not rely heavily on these information.

To ablate on the choice of VQ-VAE tokenization plus hierarchical multimodal transformers, we trained 3 additional models with alternative architectures with the same data and training objectives: (1) flat ViT: we replaced the hierarchical multimodal transformer with a single 15-layer ViT that takes in all volume tokens from all sequences in the study. Each token has a sinusoidal sequence encoding in addition to the positional encodings to indicate the sequence it belongs to, and the sequence name tokens (with sequence encoding) and the study description token are also included in the input. (2) 2D CNN: we directly encode each 2D MRI images with a 2D ResNet, and the encodings of images from a sequence is viewed as tokens from this sequence, and are fed into a hierarchical multimodal transformer architecture same as Prima. The ResNet is jointly trained during CLIP training. (3) 3D CNN: we directly encode each 3D MRI sequence with a 3D CNN architecture, and the sequence encodings from each sequence, concatenated with corresponding sequence name encodings, are fed into a $ViT_{st}$ together with study description encoding. The CNN architectures in (2) and (3) are relatively small due to GPU memory constraints (each study contains over 1000 2D MRI images on average). Prima outperformed all three ablations.

We attempted to train Prima with unsummarized long radiology reports rather than summarized short reports. We found that the model trained with long reports suffers from overfitting compared to short reports, using non-diagnostic features in MRI reports to minimize the CLIP objective. We also trained a version of Prima without patient discrimination loss. As shown in Extended Data Figure \ref{exfig:ex_data5}e, patient discrimination loss significantly accelerates model convergence. We performed an additional ablation called Simple ViT, where we train a plain ViT architecture with CLIP loss without any additional designs. We use the same architecture as the flat ViT, but without patient discrimination loss, sequence/study name, sequence dropouts, or token dropouts. The model overfits during training due to lack of inductive bias regarding the hierarchical structure of brain MRIs.

\subsection*{Explainable and trustworthy predictions}

LIME~\cite{ribeiro2016should} is a commonly used method to interpret the decision-making of black-box classification or regression models. LIME first generates a list of perturbed inputs and their corresponding masks: each perturbed input is generated by masking certain parts of the original input, and the mask simply indicates which parts are corrupted, or masked ou, and which parts are unchanged. LIME runs the black-box model on each of the masked inputs, obtaining the prediction logits for each input. LIME interpretations are obtained by fitting a linear model mapping from the masks to the logits, weighted by locality, and the weight of the linear model on each dimension of the mask indicates the importance of each masked region. LIME applies well to vision models that include an initial patching operation, similar to volume tokenization in Prima, because the masks are easily applied to the initial patches.

To determine the trusthworthiness of Prima's predictions, for example if it predicts a tumor based on the MRI's tumor regions, we run LIME on Prima's prediction across diagnostic tasks. To isolate the contribution of each sequence in an MRI study, we only input and perturb one sequence from an MRI study. Each input patch is a single 3D visual token, and the corruption process is simply token removal, such that each perturbed input only has a random subset of the visual tokens, and the mask shows which tokens are included. For our LIME experiments, 3000 masked inputs were generated and the linear model weights for each volume token were ranked and converted into color-coded visualizations as shown in Fig. \ref{fig:lime}. Qualitative evaluation was performed across all diagnostic classes (Extended Data Fig. \ref{exfig:ex_data7} and \href{www.Prima.mlins.org}{demo website}). To quantitatively evaluate Prima's selection of the volume tokens within brain tumor regions, we used the BraTS dataset that includes dense semantic segmentation masks and measured the overlap rate between the top-K LIME volume tokens and the tumor segmentation masks (Extended Data Fig. \ref{exfig:ex_data9}d). Qualitative error analysis was completed for Prima to identify suboptimal performance in specific scenarios, such as uncommon MRI sequences or protocols, is shown in Supplementary Data Fig. \ref{supfig:sup_data7}. 

\subsection*{Referral, Acuity, and Age Prediction Training}

For referral prediction tasks, we follow the same protocol as the diagnosis tasks: for $R$ total referral tasks, we freeze the CLIP-trained sequence and study encoders, and we train a 3-layer MLP that takes in the encoder output, and outputs an R-dimensional output where each dimension corresponds to the logit of a referral task (e.g. referral to pediatric neurosurgeon). The MLP is trained with positive-weighted binary cross entropy loss. For each task, we take the checkpoint of the MLP with the best performance on the held-out validation set, and save for model testing. Mappings between the radiologic diagnoses and the referrals are in Supplementary Data Table 3. For acuity prediction, the CLIP-trained encodings are frozen and we train a 3-layer MLP that takes in the encoder output and outputs a 3-dimensional vector, that corresponds to three levels of acuity: normal, medium, and high acuity. The MLP is trained with a categorical cross entropy loss on the training set. We ablated over alternative ordinal-based objectives, such as ordinal metric learning ~\cite{kondepudi2025foundation} and binary-ordinal \cite{Cheng2008-rm}. The checkpoint with the best validation performance is used for testing. Mappings between the radiologic diagnoses and the acuity classes can be found in Supplementary Data Table 3. We follow the same protocol for age prediction as above, but the 3-layer-MLP outputs a scalar value for regression using an L2 objective.

\subsection*{Prima on Public Datasets}

To further evaluate the generalizability and transferability, we evaluate Prima on several publicly available datasets. We divide the datasets into 2 groups. The group 1 aims to test generalizability and includes MRI datasets with one or more of the study diagnoses, namely in-domain. The group 2 aims to test transferability and includes MRIs with diagnoses outside of our study diagnoses, namely out-of-domain. These are out-of-domain because CLIP training with the radiology provides no supervision for these tasks. For each dataset in the first group, we directly use Prima and the diagnosis-specific MLP head to predict on MRI studies, and report true positive rate as the evaluation metric. The datasets we include in group 1 are BRATS 2021 (adult glioma) \cite{baid2021rsna}, NYUMets (metastasis) \cite{oermann2023longitudinal}, UCSF-BMSR (metastasis) \cite{rudie2024university}, and Stroke (large vessel stroke) \cite{liu2023large}. The group 2 includes datasets for autism spectrum disorder (ABIDE \cite{di2014autism}) and dementia (ADNI \cite{petersen2010alzheimer} and OASIS-1 \cite{marcus2007open}). For each task in each dataset, we obtain a single embedding vector for the MRI studies. ABIDE has one sequence per study and we directly encode ABIDE sequences with $ViT_{seq}$. Otherwise, we use the full Prima encoder. We then trained a 2-layer MLP that outputs a single logit for the task and trained with a binary cross entropy loss. We perform a 5-way cross-validation on each task, and report average validation performance and standard deviation in Fig. \ref{fig:results}. Results were compared to recent baselines for ABIDE and ADNI tasks \cite{barbano2024anatomical}, and for OASIS-1 tasks \cite{saratxaga2021mri}.

\subsection*{VLM zero-shot benchmarking details}

We evaluated the performance of several state-of-the-art, pre-trained VLMs as zero-shot baselines on the radiologic diagnosis tasks. We focus on three categories of publicly-available pre-trained VLMs: CLIP family (CLIP-base~\cite{Radford2021-wf}, CLIP-large~\cite{Radford2021-wf}, PubMedCLIP~\cite{eslami2023pubmedclip}, BioMedCLIP~\cite{zhang2023biomedclip}, Blip-2~\cite{li2023blip2bootstrappinglanguageimagepretraining, chen2024medblip}), Llava family (Llava-1.5-7B~\cite{liu2023improvedllava}, Llava mistral, Llava-Med~\cite{li2023llavamed}, Llava-3D~\cite{zhu2025llava3dsimpleeffectivepathway}), and MedFlamingo~\cite{moor2023medflamingo}. Other relevant neuroimaging models were excluded due to lack of open-source implementation or publicly releasesd model \cite{wood2024self}. We randomly sample a subset of the clinical test set to evaluate the VLMs. For a diagnosis task with $p$ positives in the testing set, we randomly sample $\min(p,100)$ positive samples and $\min(p,100)$ negative samples from the testing set to form a balanced subset. 

Since the models only support 2D image inputs, we devised the following procedure to obtain diagnosis predictions: for each diagnosis, we asked a domain expert (T.C.H.) to manually choose the most diagnostic sequences (T1W, T2W, FLAIR, DWI, etc.) for each diagnosis. Then, for each MRI study in the balanced subset, we select the diagnostic sequence $S$. We obtain a relative prediction score for the diagnosis using $S$ as follows:

\begin{itemize}
    \item \textbf{CLIP family}: For each 2D image in $S$, we compute the cosine similarity between the 2D slice embeddings of the image and a prompt (e.g. "A Brain MRI scan of a patient with $<$diagnosis-name$>$"), and we take the maximum similarity over all slices in $S$, since only a portion of the images in $S$ may contain relevant information about the diagnosis. We tried several different prompts and reported the performance of the best-performing prompt for each task.
    \item \textbf{LLaVa family}: This family only supports one image input per query. We input each 2D slice in $S$ into the model together with a prompt (e.g. "Does this image show an MRI scan with $<$diagnosis-name$>$?") We compute a prediction score with normalized yes-no logits, that is $\frac{p(yes)}{p(yes)+p(no)}$, where $p(yes)$ and $p(no)$ indicates the prediction probability of the first output token being "yes" or "no", respectively. We take the max score over all images in $S$ as the final prediction score for the study. We tried several different prompts and reported the performance of the best-performing prompt for each task.
    \item \textbf{Med-Flamingo}: This model supports multiple images per input query. We input all 2D images in $S$ interleaved with the text prompts "This is image number 1 in a Brain MRI scan", "This is image number 2 in a Brain MRI scan", etc. Then, we ask the question "Does this patient have $<$diagnosis-name$>$?", and measure the prediction score with normalized yes-no logits as above. We also experiment an additional setting where we input the sequence and study names of each 3D scan before the images, i.e. "You are provided with 2D slices of a 3D brain MRI scan (Sequence $<$sequence\_name$>$ from study $<$study\_name$>$)."
\end{itemize}

The higher the VLM's prediction score, the more positive the model's indication for the diagnosis. Therefore, we can compute ROC curve over the prediction scores. We report the performance of the best-performing model on each task within each model family together with Prima's zero-shot performance on the same balanced subsets in Fig. \ref{fig:results}b, with task-wise comparison in Extended Data Fig. \ref{exfig:ex_data5}a. We include best-performing prompts for each model for each task in Supplemental Data Table 6. We could not report performance of OpenAI's GPT VLMs because they refuse to give diagnostic predictions for MRI images.

\subsection*{MLP Probing of 2D CLIP models}

In addition to zero-shot evaluation, we compared the results of MLP-probing of the 2D CLIP models, from "CLIP family" above, over the UM-220K dataset to Prima's MLP-probing performance. For each CLIP model, the 2D vision encoder was used to encode each MRI slice. We then performed average pooling over the encodings to get the final study representation. Then, we fit the same 3-layer MLP classification head over UM-220K training data in the same manner as Prima. We report classification performance on testing data. The results are shown in Extended Data Figure~\ref{exfig:ex_data5}b. The 2D CLIP models show improved performance when trained on UM-220K dataset compared to zero-shot performance, indicating the value of health system-scale training. Baseline CLIP models underperform Prima due to the hierarchical vision transformer design and 3D modeling.

\subsection*{Prima with clinical context}

Radiologists use clinical context to better interpret MRI studies. Similarly, we hypothesized that Prima would benefit from having the clinical context when predicting radiologic diagnosis. To perform this evaluation, we first use GPT-4o-mini to parse the clinical history and indications from each patient's radiology report. We then use OpenAI's text-embedding-3-small model to obtain 1536-dimensional embeddings for each patient's clinical context. Then, we concatenate each clinical embedding with the 10240-dimensional MRI-study embedding from Prima for each study, forming a 11776-dimensional embedding for each study. We then performed the MLP-probing over 52 diagnostic tasks, and reported performance in Figure~\ref{fig:results}a.

\subsection*{Bias and Fairness criteria}
We used odds ratios to measure relative disparity between a reference group and a sensitive group to evaluate potential systemic biases. A two-sided Fisher exact test was used to compute p-values with multiple hypothesis correction. Null hypothesis was that the odds of longer turnaround time was not higher in the sensitive group compared to the reference group.

To evaluate the fairness of Prima, we used the equalized odds framework combined with intersectional and diagnostic subgroup analysis \cite{Hardt2016-fa, Barocas2023-nn}. Algorithmic fairness under this framework uses a separation criterion 
defined over a sensitive attribute $A \in \{a,b\}$, a classifier $\hat{Y}$, and target variable $Y$, such that $\hat{Y} \perp A \mid Y$. The conditional independence states that the odds of predicting `positive' or `negative' are independent of the sensitive attribute conditioned on the target variable. The separation criteria for true positive rate (TPR), $\mathbb{P}\{\hat{Y} = 1 \mid Y = 1\}$, and false positive rate (FPR), $\mathbb{P}\{\hat{Y} = 0 \mid Y = 0\}$ are, respectively: 
\begin{equation}
\mathbb{P}\{\hat{Y} = 1 \mid Y = 1, A = a\} = \mathbb{P}\{\hat{Y} = 1 \mid Y = 1, A = b\}
\end{equation}
\vspace{-10pt}
\begin{equation}
\mathbb{P}\{\hat{Y} = 0 \mid Y = 0, A = a\} = \mathbb{P}\{\hat{Y} = 0 \mid Y = 0, A = b\}
\end{equation}
Satisfying both criteria demonstrates equalized odds. Intersectional analysis includes combining sensitive attributes, such that equalized odds are computed over the intersection of patient sets defined by individual attributes, for example, black females or Asians living in southwest Michigan. Finally, we evaluate these metrics for specific diagnoses, such as neoplastic lesions, ischemic strokes, and pediatric diagnoses. Complete algorithmic fairness experiments can be found in Extended Data Fig. \ref{exfig:ex_data10} and Supplementary Data Fig. 6. As Prima is well-suited for AI screening and decision support tool, TPR is the most important metric to evaluate and is called equalized opportunity as a fairness metric \cite{Vaidya2024-dw}; missing diagnoses are worse than predicting them when the patient is normal. Equalized opportunity evaluates if a classifier's TPR is equalized across sensitive attributes for model fairness and nondiscrimination. TPR was calculated for a sensitive subgroup, $G$, as:
\begin{equation}
\text{TPR}_{G} = \frac{\sum_{i \in G} \mathbbm{1}(y_i = 1 \land \hat{y}_i = 1)}{\sum_{i \in G} \mathbbm{1}(y_i = 1)},
\end{equation}
and for the full study population as
\begin{equation}
\text{TPR}_{Pop.} = \frac{\sum_{i=1}^{N} \mathbbm{1}(y_i = 1 \land \hat{y}_i = 1)}{\sum_{i=1}^{N} \mathbbm{1}(y_i = 1)}.
\end{equation}

Deviations from equalized opportunity were measured using TPR disparity:
\begin{equation}
\text{TPR disparity} = \text{TPR}_{G} - \text{TPR}_{Pop.} \in [-1, 1]
\end{equation}
with larger absolute values, $\vert \text{TRP disparity} \vert$, reflecting greater algorithmic bias. Examples of sensitive attributes that defined subgroups in our study were population density, geography, scheduling, race, ethnicity, sex, age, etc. 


\subsection*{Equalized opportunity experiment design}
For each fairness experiment, we used bootstrap sampling to estimate the sensitive subgroup TPR and the population TPR \cite{Vaidya2024-dw}. We randomly sample 200 patients with replacement from a subgroup and 200 diagnosis-matched patients from the study population with replacement. We then computed the TPR values for the bootstrap sample subgroup and the population. The process was repeated for 20 iterations. We performed P-value testing using a one-sided, non-parametric Mann–Whitney U statistical test. Our null hypothesis was that the subgroup TPR distribution was not less than the population TPR distribution. Lower TPR rates for the sensitive subgroups represent bias that causes harm through more false negatives. 

\subsection*{Computational hardware and software}

All training for Prima and ablations were performed on high-performance computing clusters hosted by the University of Michigan Advanced Research Computing (ARC) group. Specifically, all computation was done on 6 nodes on the HIPAA-compliant Armis2 clusters of ARC. Each of the 6 nodes contain 8 NVIDIA L40S GPUs (48GB DRAM each), 64 CPU cores (Intel Xeon Platinum 8358), and 503G of RAM.
 
The training and inference programs are written in Python with PyTorch library. Flash-attention~\cite{dao2022flashattentionfastmemoryefficientexact} was used for reduced GPU memory usage and accelerated training. HuggingFace Transformers library~\cite{huggingfacetransformers} was used to access and run large pre-trained models, including gpt2, CLIP-base (openai/clip-vit-base-patch32), CLIP-large (openai/clip-vit-large-patch14), PubMedCLIP (flaviagiammarino/pubmed-clip-vit-base-patch32), BiomedCLIP (microsoft/BiomedCLIP-PubMedBERT\_256-vit\_base\_patch16\_224), Llava-1.5-7B (llava-hf/llava-1.5-7b-hf), Llava-mistral (llava-hf/llava-v1.6-mistral-7b-hf), Llava-Med (microsoft/llava-med-v1.5-mistral-7b), and MedFlamingo (med-flamingo/med-flamingo). Our ViT implementation is adapted from \url{https://github.com/lucidrains/vit-pytorch}.

Training Prima requires 1 HPC node. The training is distributed amongst 8 GPUs with DataParallel, and it took approximately 50 days to train our final Prima model from scratch. Pre-training the GPT-2 language model and the sequence name encoder each took about one day. Although each model is trained with only one node, having multiple nodes allowed us to experiment with various design choices, configurations and ablations in parallel. Inference throughput of our model is efficient: inferencing all 29,438 testing studies with 1 L40S GPU and 8 CPU cores takes about 75 minutes (over 6.5 studies per second).

\section*{Ethics and inclusion statement}
Our research was approved by the University of Michigan Institutional Review Board (HUM00229133). All MRI data was acquired under secondary data usage. The methods were carried out in accordance with the IRB’s guidelines, regulations, and policies. All human subjects that met inclusion criteria as stated above were included in the study. 

\section*{Data availability}
The Prima model parameters will be publicly available for investigational use only under an MIT license. Institutional Review Board approval was obtained from University of Michigan for MRI data collection. Restrictions apply to the availability of raw patient MRI imaging data, which were used with institutional permission through IRB approval for the current study, and are thus not publicly available. All data sharing between medical centers is regulated through data use agreements with the study authors. A similar data-sharing protocol may be established for interested investigators. Please contact the corresponding author (T.C.H.) for any requests for data sharing. All requests will be evaluated based on institutional and departmental policies to determine whether the data requested is subject to intellectual property or patient privacy obligations. Data can only be shared for non-commercial academic and investigational purposes.

\section*{Code availability}
All code was implemented in Python (version 3.9) using PyTorch (2.3.1) and Transformers (4.37.0) as the primary machine learning framework. All code and scripts to reproduce the training and inference of Prima are available on GitHub at \href{https://github.com/MLNeurosurg/Prima}{MLNeurosurg/Prima} under an MIT license. 

\section*{Acknowledgments}
We would like to thank Karen Eddy, Gary Laderach, Brock Palen for providing technical support. David Hanauer for support with the University of Michigan Electronic Medical Record Search Engine (EMERSE). Anthony Rosenzweig for scientific guidance.

This work was supported by the following National Institute of Health (NIH) funding sources: K12NS080223 (T.C.H.). This work was supported by the Chan Zuckerberg Initiative (CZI), Frankel Institute for Heart and Brain Health (T.C.H.), the Mark Trauner Brain Research Fund, the Zenkel Family Foundation (T.C.H.), Ian’s Friends Foundation (T.C.H.) and the UM Precision Health Investigators Awards grant program (T.C.H.).

This research was also supported, in part, through computational resources and services provided by Advanced Research Computing, a division of Information and Technology Services at the University of Michigan.

\bibliographystyle{unsrt}
\bibliography{paperpile_rev,yiwei_additional}

\clearpage
\appendix

\section{Extended Data Figures}
\begin{enumerate}
    \item Expanded Workflow and Prima Architecture
    \item Overview of UM-220K Dataset
    \item LLM annotations and neuroradiology language models
    \item MRI volume tokenization
    \item MRI-report contrastive pre-training
    \item Extended Prima diagnostic performance results
    \item Diverse clinical examples of Prima explainibility
    \item Multi-label analysis of Prima
    \item Extended clinical task and transfer learning results
    \item Intersectional and equalized odds analysis
\end{enumerate}

\section{Supplementary Data}
\subsection{Supplementary Data Figures}
\begin{enumerate}
    \item SQL query for UM-220K search
    \item GPT-3.5 prompt for report summarization with summarized report examples
    \item GPT-4 prompt for report classification
    \item PyTorch-style pseudocode for VQ-VAE with random permutations
    \item PyTorch-style pseudocode for Prima MRI-Report CLIP
    \item Subgroup performance and fairness analysis
    \item Error analysis with LIME visualizations
\end{enumerate}

\subsection{Supplementary Data Tables}
\begin{enumerate}
    \item Demographic information for UM-220K dataset
    \item Demographic information for prospective testing dataset
    \item Diagnosis-Referral-Priority mapping
    \item Demographic summary of retrospective fairness dataset
    \item Demographic summary of prospective fairness dataset
    \item Zero-shot prompts
    \item Comprehensive classification metrics for Prima
    \item Label and group equalized odds analysis
\end{enumerate}

\clearpage
\renewcommand{\figurename}{Extended Data Figure}
\setcounter{figure}{0}

\begin{figure*}[p!]
    \centering\includegraphics[scale=0.78]{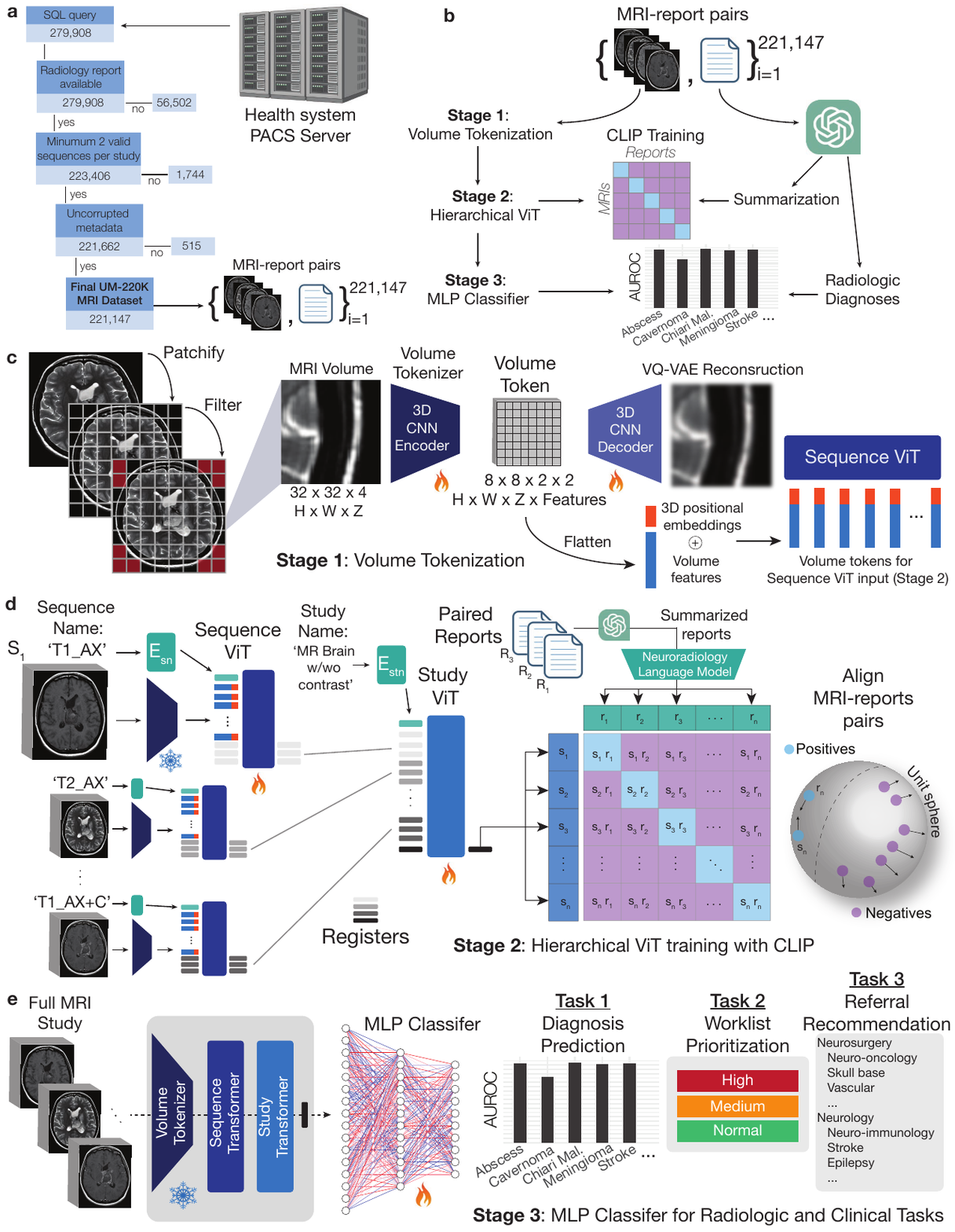}
    \caption{\textbf{Expanded Workflow and Prima Architecture}. Caption on next page.}
    \label{exfig:ex_data1}
\end{figure*}

\begin{figure*}[p!]
  \contcaption{\textbf{Expanded Workflow and Prima Architecture} \textbf{a}, Our health system Sectra PACS server was queried for all cranial MRIs. We then filtered these MRIs based on the availability of an associated radiology report and having a minimum of 2 series per study. We then ensured that all metadata was present, resulting in a total of 221,147 UM-220K dataset. \textbf{b}, Overview of the stages of training Prima on UM-220K, which includes volume tokenization, hierarchical ViT training with CLIP objective function, and transfer learning to predict radiologic diagnoses. An LLM provides radiology report summarization and diagnostic labels for reliable, accurate, and scalable vision-language modeling. \textbf{c}, Volume tokenization stage involves dividing each MRI volume into smaller subvolume patches of shape 32x32x4, removing background tokens, and encoding each subvolume using a VQ-VAE encoder. The latent VQ-VAE tokens are then passed forward to the sequence ViT with the concatenated positional encodings. \textbf{d}, The hierarchical ViT is trained using a CLIP objective on frozen volume token features. The sequence ViT is a multimodal transformer that takes as input both the volume tokens and the embedded free-text sequence description. The series registers are passed forward to the study ViT that outputs a single representation for the full MRI study. The paired reports are summarized and passed through a pre-trained neuroradiology model to align the MRI study and the paired report. \textbf{e}, A transfer learning strategy is used such that the volume tokenizer, sequence and study transformers are frozen, and an MLP is trained on the learned study features for radiologic and clinical task prediction.}
\end{figure*}

\clearpage
\begin{figure*}[p!]
    \centering\includegraphics[scale=0.78]{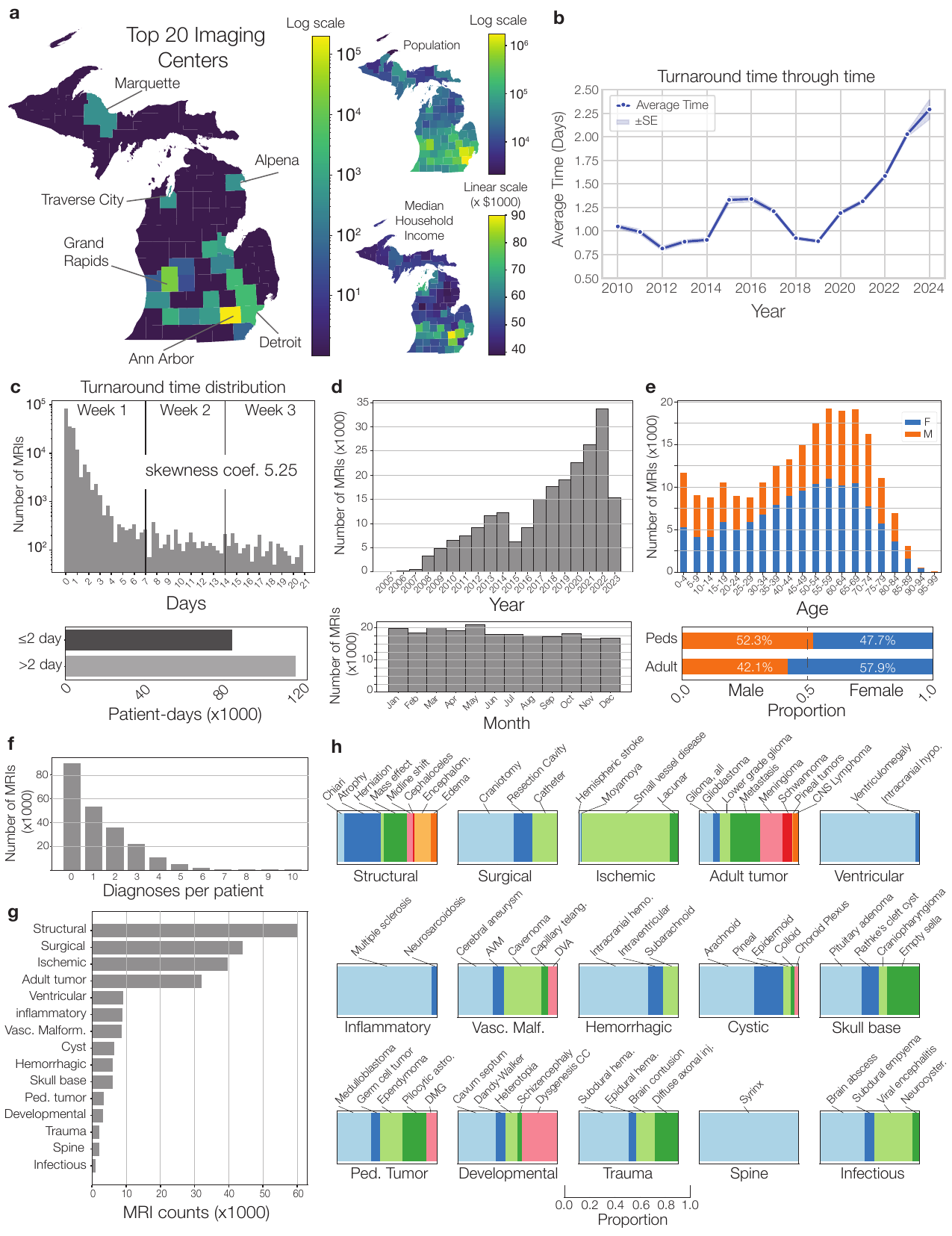}
    \caption{\textbf{Overview of UM-220K Dataset}. Caption on next page.}
    \label{exfig:ex_data2}
\end{figure*}

\begin{figure*}[p!]
\contcaption{\textbf{Overview of UM-220K Dataset}. Descriptive details of the UM-220K dataset are shown here. \textbf{a}, The majority of MRI data is collected from patient's with permanent residence in the state of Michigan. A geospatial map shows the counties with the top 20 imaging centers across the state. The top centers are located in populous regions and regions with higher median income based on census data (www.census.gov). The Upper Peninsula and Northern Michigan are lower resource settings, rural areas, and most susceptible to experiencing longer turnaround times (Fig. \ref{fig:bias}). \textbf{b}, Average turnaround time through time. We observed a steady year-to-year increase in turnaround time since 2019. This increase correlates with increasing MRI demand and imaging volumes at our health system. \textbf{c}, The distribution of turnaround times. The distribution shows a severe right skew, with a Fisher-Pearson coefficient of skewness of 5.25. The majority of turnaround time measured in patient-days is in the right tail distribution greater than 2 days. These results prompted us to target turnaround time as a metric of algorithmic fairness. \textbf{d}, Distribution of MRI counts through time and divided by month of year. We observed a consistent increase in the number of MRIs/year.  \textbf{e}, Age and sex distribution of the UM-220K. \textbf{f}, Distribution of patients by the number of diagnoses per patient, including patients with no diagnoses. \textbf{g}, Distribution of diagnostic categories and (\textbf{h}) the granular radiology diagnoses for for each category. The aims was to have a broad set of diagnostic categories that spanned the full diagnostic spectra and to include clinically meaningful and actionable diagnostic classes. 
  }
\end{figure*}

\clearpage
\begin{figure*}[p!]
    \centering\includegraphics[scale=0.77]{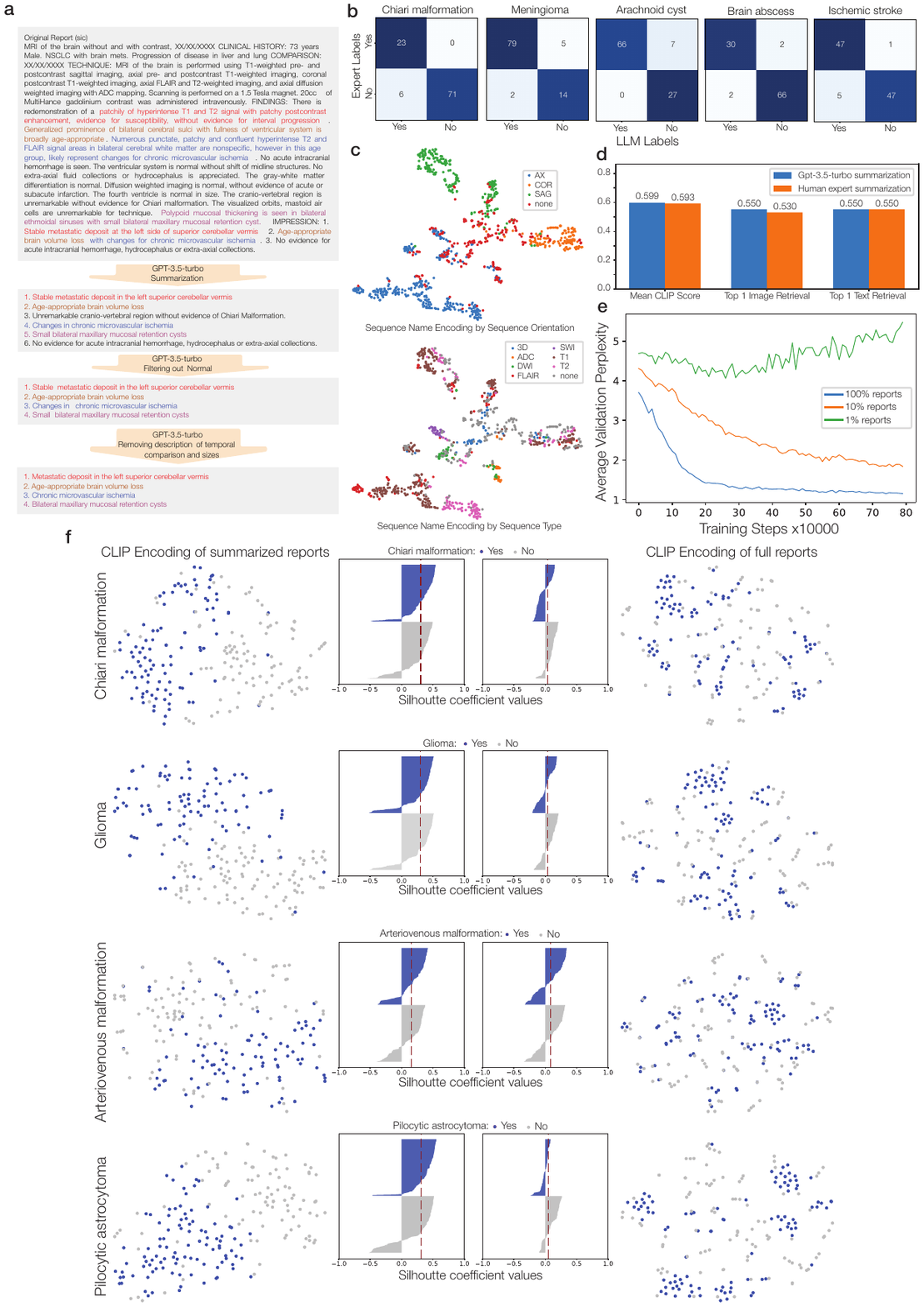}
    \caption{\textbf{LLM annotations and neuroradiology language models}. Caption on next page.}
    \label{exfig:ex_data3}
\end{figure*}

\begin{figure*}[p!]
    \contcaption{\textbf{LLM annotations and neuroradiology language models}. \textbf{a}, Example of an original MRI report with highlighting of the major findings in the MRI study. GPT-3.5-turbo is then prompted in stages to summarized the report to achieve an itemized report summary of positive findings. Additional details regarding prompting are in Supplementary Data Fig. 2. \textbf{b}, The report classification performance of GPT-4 for 5 diagnoses from the different diagnostic categories: structural, adult tumor, cystic lesions, infectious, and vascular ischemic. Prompting details of GPT-4 classification are in Supplementary Data Fig 3.  \textbf{c}, tSNE visualizations of the sequence name encoding after training the sequence name encoder, $E_{sn}$, using a CLIP objective. The model correctly encodes sequence names based on both imaging planes and sequence type. This effectively prompts $ViT_{seq}$ for better MRI sequence feature extraction. \textbf{d}, Comparison on Prima CLIP-score and retrieval performance between GPT-3.5-turbo summarized reports and human expert summarized reports on 100 prospective studies. We observe no statistically significant difference between the two summarizations on CLIP scores (P=0.62, paired two-sided t-test) or retrieval metrics (P>0.80, McNemar's test), indicating that Prima, although trained on GPT-3.5-turbo summarized reports, is aligned with radiologist summarization. \textbf{e}, Line plot shows the average validation perplexity during training of the neuroradiology language model given different percentages of the report data. Neuroradiology language model was trained using next-word prediction and benefits from increased training data, approaching the lower perplexity bound of 1 with 100\% of the MRI reports. \textbf{f}, tSNE plots of the CLIP encoded summarized reports versus full reports. The summarized reports show better label-conditional clustering compared to the full reports. These results demonstrate the challenge of using full reports for CLIP objective, which contain extraneous or non-informative textual details that can degrade visual representation learning. The center panels are Silhouette plots to quantify cluster quality. The red dotted line is the average Silhouette coefficient. Larger values represent better clustering results.}
\end{figure*}

\clearpage
\begin{figure*}[p!]
    \centering\includegraphics[scale=0.8]{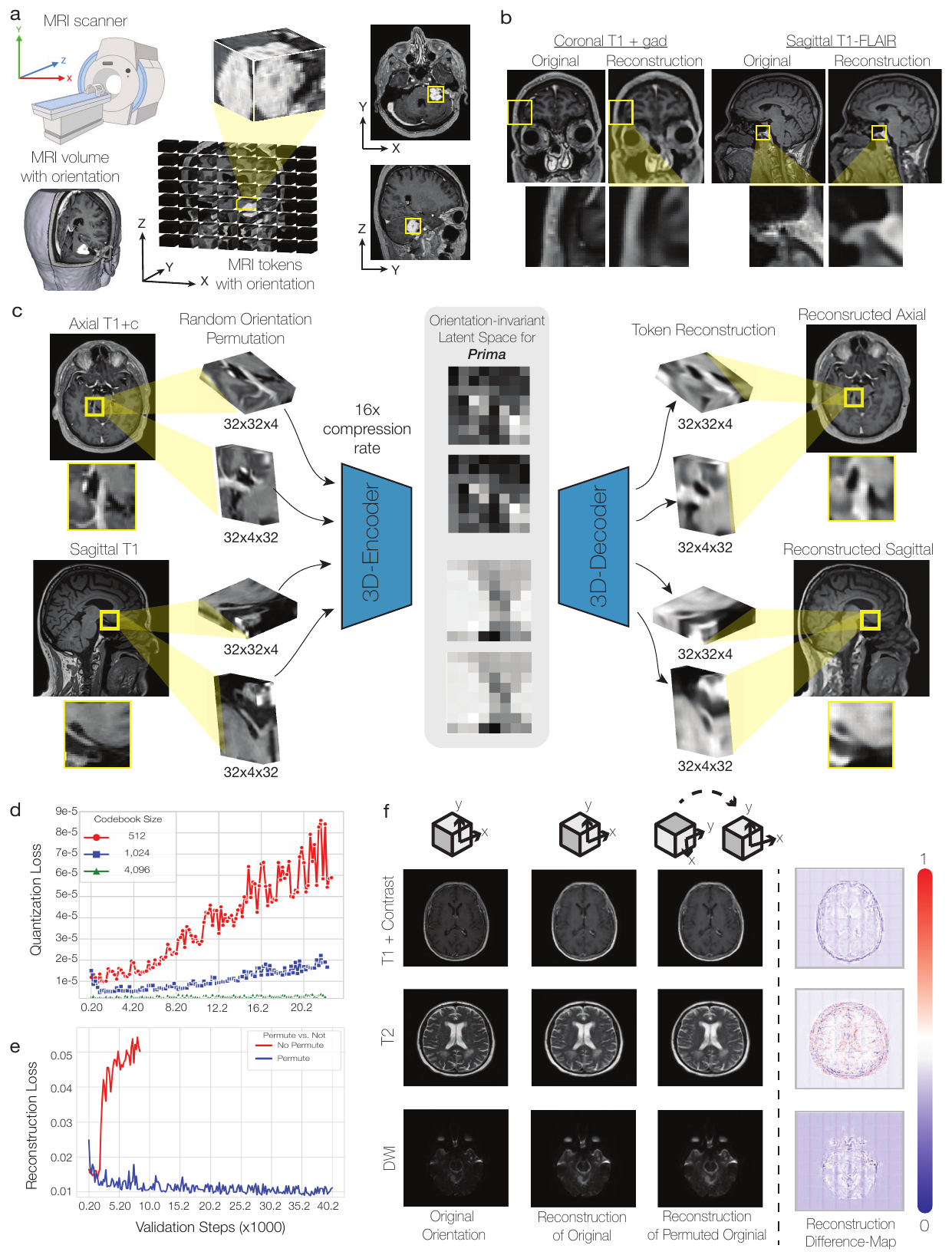}
    \caption{\textbf{MRI volume tokenization}. Caption on next page.}
    \label{exfig:ex_data4}
\end{figure*}

\begin{figure*}[p!]
    \contcaption{\textbf{MRI volume tokenization} \textbf{a}, MRI scanners acquire images with specified orientations (e.g. LAS, RAS, etc) and planes (e.g. axial, coronal, sagittal). The MRI tokens will have the same orientation and plane as the source MRI sequence after patching. \textbf{b}, Examples of VQ-VAE reconstructions in different MRI sequences and orientations. \textbf{c}, Because Prima takes as input multiple different orientations and imaging planes, the volume tokenizer should be orientation invariant, meaning the representation of the same anatomic region should not change if imaged in axial versus coronal plane or LAS versus RAS orientation, for example. We used two strategies: random orientation permutations and 3D-CNN encoders. Our VQ-VAE volume tokenizer is encouraged to encode each volume token equivalently across all orientations under a reconstruction loss. Examples of MRI subvolumes are shown in different orientations after permutation. The latent volume tokens with near-equivalent latent encodings are shown in the center panel. with the reconstructions after the decoder on the right.  \textbf{d}, Ablation study over the codebook sizes shows the quantization loss validation curves. Larger codebook sizes led to less overfitting and better reconstructions. \textbf{e}, Reconstruction losses with and without random orientation permutations. Random permutations regularized the VQ-VAE and resulted in higher-quality reconstructions and lower reconstruction losses. \textbf{f}, Examples of reconstructions before and after orientation permutations for different MRI sequences. Reconstructions are perceptually equivalent after forward pass through the VQ-VAE model regardless of orientation or imaging plane. Subtle reconstruction differences can be seen on difference maps.}
\end{figure*}

\clearpage
\begin{figure*}[p!]
    \centering\includegraphics[width=\textwidth]{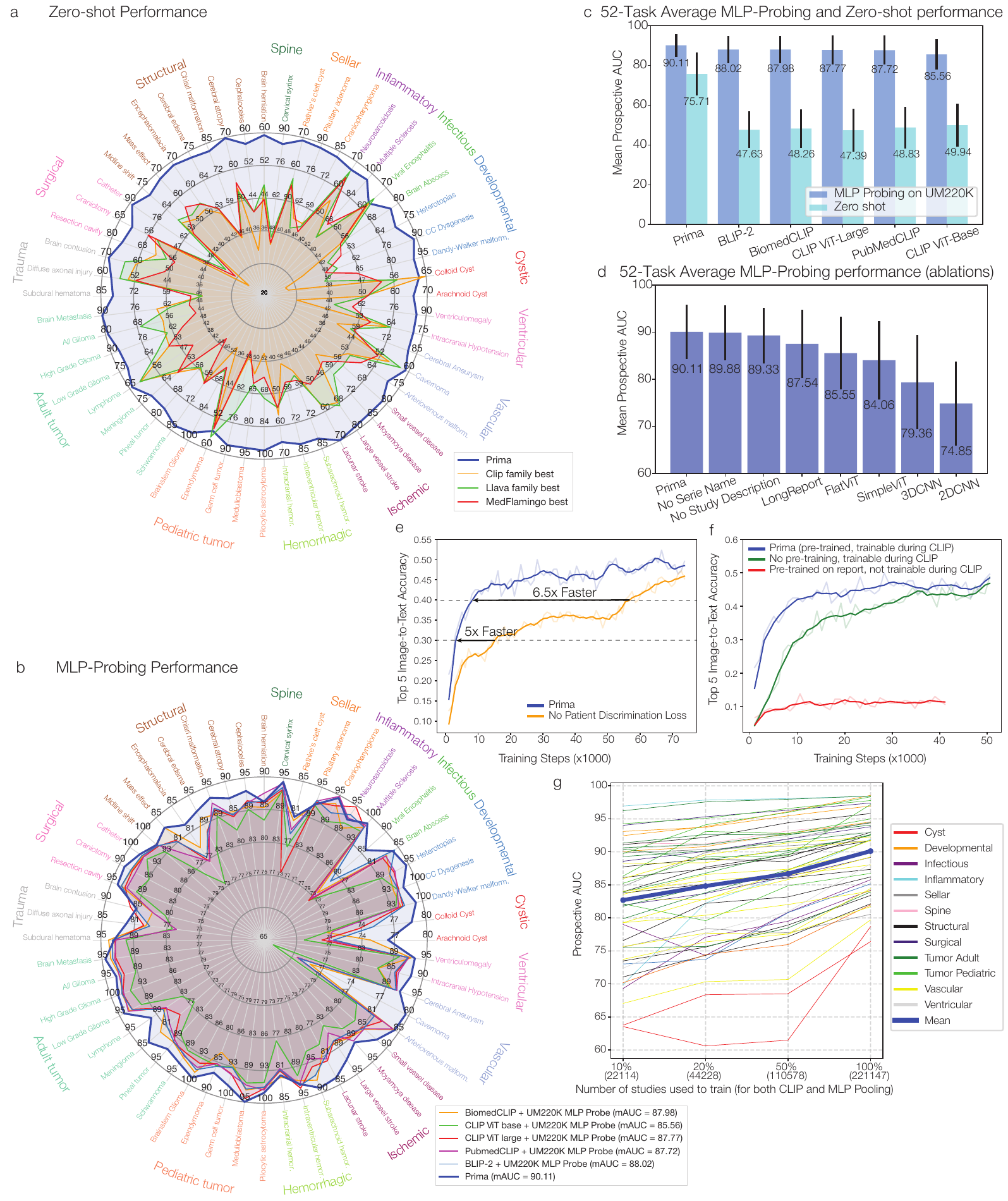}
    \caption{\textbf{MRI-report contrastive pre-training and performance}. Caption on next page.}
    \label{exfig:ex_data5}
\end{figure*}

\begin{figure*}[p!]
    \contcaption{\textbf{MRI-report contrastive pretraining}. \textbf{a}, Radar plot of prospective zero-shot performance (in AUC) comparison between Prima and open-source pre-trained VLM families across 52 tasks. The baseline performances reported for each VLM family is from the best-performing model in the family for each task. Zero-shot Prima significantly outperforms all baselines across the majority of tasks.  \textbf{b}, Radar plot of prospective MLP-probing performance on UM-220K across 52 tasks, between Prima MRI study representations and average-pooled study representations from open-source pre-trained 2D CLIP-like models, all probed over UM-220K training data. Prima outperforms the 2D CLIP models, highlighting the importance of CLIP pre-training on 3D volumes and whole studies. \textbf{c}. Bar plot of Prima performance against baselines zero-shot performance and MLP-probing performance after training a diagnostic classifier on UM-220K data. \textbf{d}, Mean prospective AUC across model design choices. Prima architecture outperformed other model designs.  \textbf{e}, Top 5 validation set image-to-text retrieval accuracy of Prima with and without the patient discrimination loss. We see over a 5x speed-up in training time when using the patient discrimination loss. \textbf{f}, Ablation experiments over the neuroradiology language model. We found that pre-training on radiology reports resulted in more efficient training and that updating the language model during CLIP training was essential for MRI representation learning. \textbf{g}, AUC performance of Prima on each of the 52 tasks with various amounts of training data for both CLIP training and MLP probing. We have not observed an upper bound on performance and the data provides evidence that additional MRI data will continue to improve performance, even at fixed model capacity and compute budget \cite{Kaplan2020-ia}. These results emphasize the importance of health system-scale training and the health system-as-data engine framework.}
\end{figure*}
\clearpage

\begin{figure*}[p!]
    \centering\includegraphics[width=\textwidth]{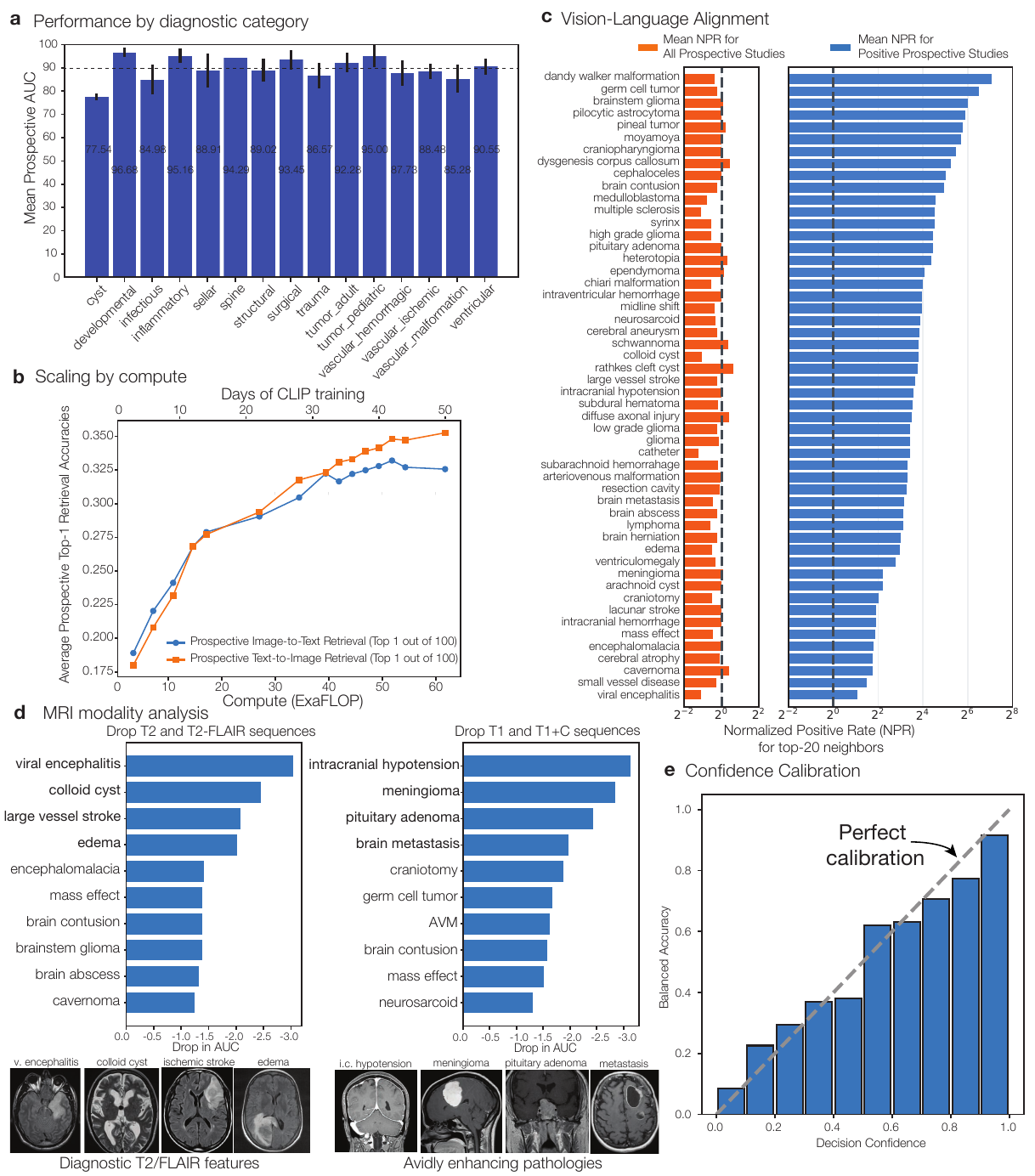}
    \caption{\textbf{Extended Prima performance results.} Caption on next page.}
    \label{exfig:ex_data6}
\end{figure*}

\begin{figure*}[p!]
    \contcaption{\textbf{Extended Prima performance results} 
    \textbf{a}, Performance in each diagnostic category averaged over the diagnostic tasks. \textbf{b}, Average Top-1 retrieval accuracy on the prospective testing cohort by days of CLIP training and amount of computation in ExaFLOPs. We see a continued, unbounded increase in performance after 50 training days. \textbf{c}, We show the normalized positive rate (NPR) of the top 20 nearest retrospective neighbors for prospective instances on each prediction task. NPR is calculated by $\frac{\text{average positive rate in top 20 nearest retrospective neighbors}}{\text{positive rate in retrospective set}}$. In other words, for any prospective MRI study with a set of positive labels, the NPR indicates how many times as likely the top 20 nearest neighbors are to have same label, compared to the positive rate of the full dataset. For each diagnostic task, we compute NPR averaged across all prospective studies (\textcolor{orange}{orange}) and across positive prospective studies only (\textcolor{blue}{blue}). For example, for Dandy-Walker Malformation (DWM), the prospective positives have an average NPR of 133.91, indicating that the average positive rate of top-20 nearest neighbors of all prospective instances with DWM (4.057 out of 20, around 20\%) is 133.91 times the overall DWM positive rate (335 out of 221147, around 0.151\%). The expected NPR value for randomly distributed examples is 1 (dotted line). The mean NPR for the positive prospective instances (blue bars) exceeds 1 by a large margin. The orange bars are around or less than 1, indicating that the neighbors of positive studies are much more likely to be positive for the same prediction. Therefore, this analysis demonstrates that Prima embeddings tend to group studies with the same diagnoses closer together. \textbf{d}, We investigate Prima's use of MRI sequence modalities for radiologic diagnosis by dropping T1-weighted and T2-weighted modalities during inference, and showed the top 10 diagnoses with the highest drop in performance. For diagnoses that produce T2 hyperintensity such as encephalitis, acute ischemia, and cerebral edema, Prima shows an expected drop in diagnostic performance, indicating Prima is using known radiologic features to diagnose specific pathologies. Even characteristically T2 hypointense lesions, such as colloid cysts, are detected by Prima. Similarly, Prima performance expectedly drop when diagnosing avidly enhancing pathology in the absence of T1 and T1-contrasted modalities. These modality-dropping experiments demonstrate that Prima is utilizing MRI sequences for radiologic diagnosis, concordant with human interpretation. \textbf{e}, A reliability diagram shows that Prima's diagnostic outputs are well calibration~\cite{guo2017calibrationmodernneuralnetworks}. The sigmoid prediction logit is the confidence score. We then calculate the balanced accuracy of the predictions with a confidence within each 0.1-binned interval. A calibrated classifier should have a confidence score that matches the accuracy within each interval (gray dashed line).}
\end{figure*}
\clearpage
\begin{figure*}[p!]
    \centering\includegraphics[width=\textwidth]{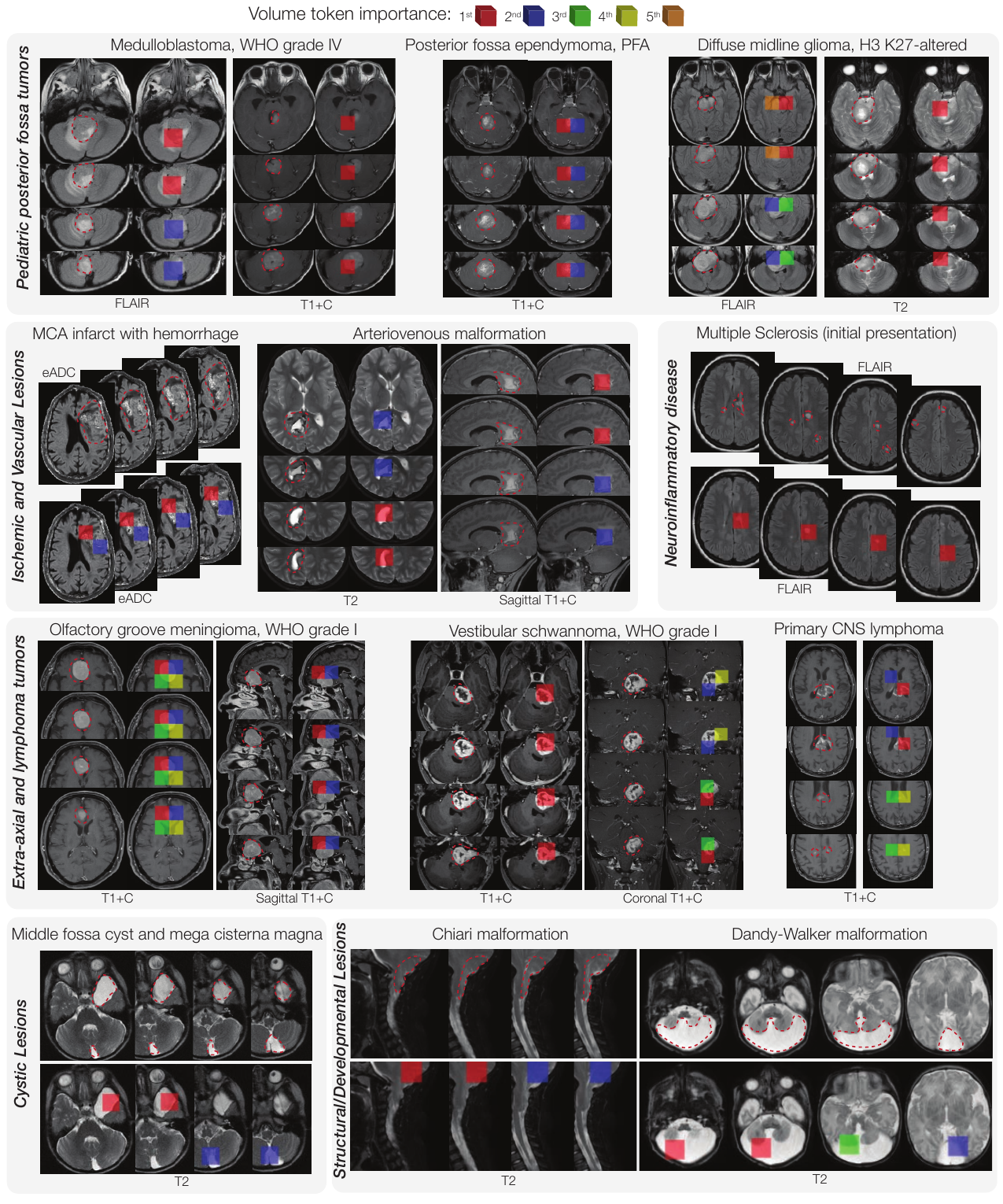}
    \caption{\textbf{Diverse Clinical Examples of Prima Explainability}. Caption on next page.}
    \label{exfig:ex_data7}
\end{figure*}

\begin{figure*}[p!]
    \contcaption{\textbf{Diverse Clinical Examples of Prima Explainibility.} We show a diverse selection of patients from our prospective testing cohort with associated LIME importance scores on the volume tokens. The top 5 tokens are color-coded according to the legend above. Prima correctly identifies the pathologic regions in all the clinical scenarios presented above, including pediatric posterior fossa tumors, vascular malformations, ischemic lesions, adult brain tumors, brain cysts, and developmental abnormalities. High LIME score tokens that localize to the pathologic regions demonstrate trustworthy Prima predictions. Red dashed line represented expert annotated pathologic regions. Full interactive demonstration with LIME visualizations and predictions can be found at \href{https://prima.mlins.org/}{prima.mlins.org}.}
\end{figure*}

\clearpage
\begin{figure*}[p!]
    \centering\includegraphics[scale=0.80]{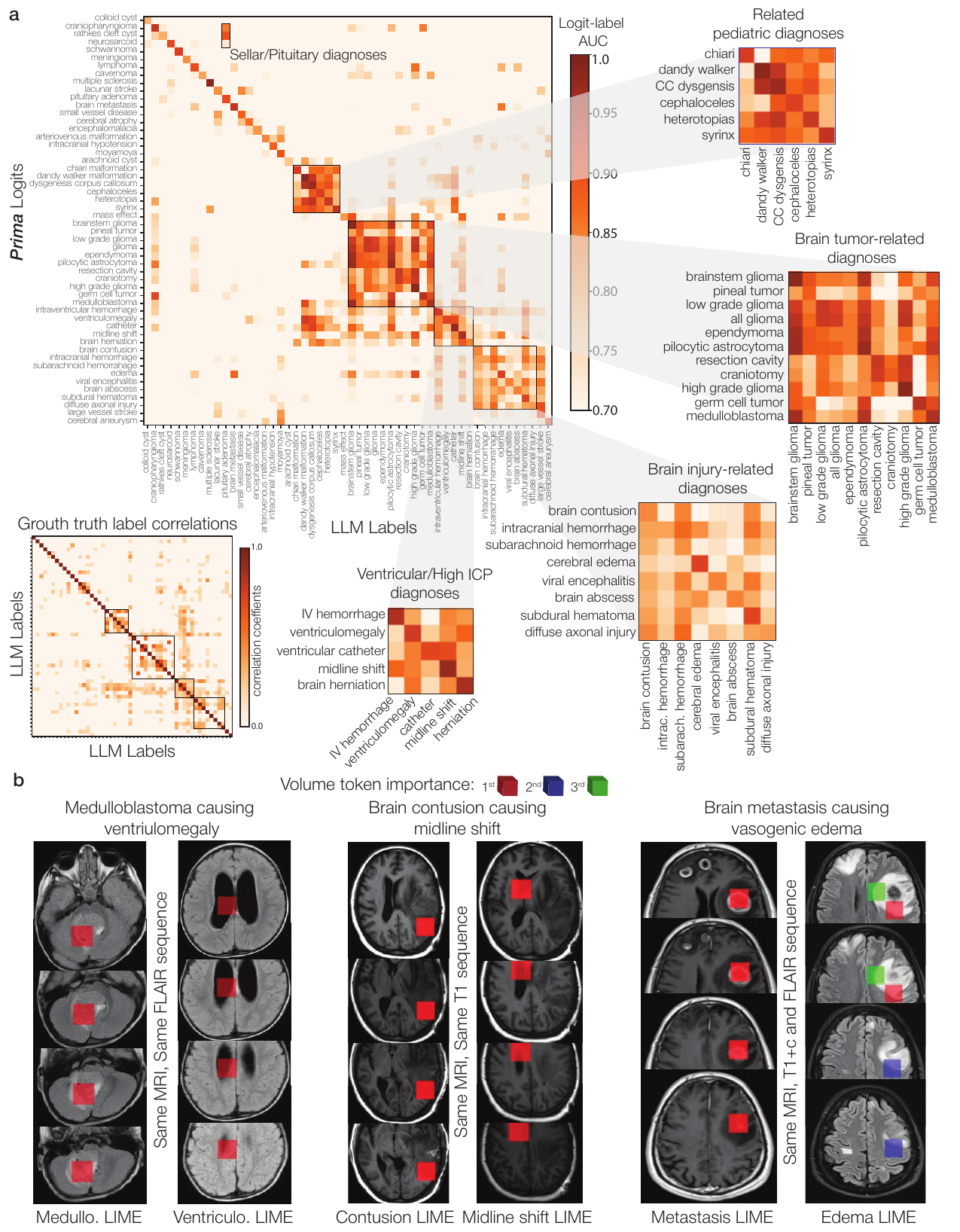}
    \caption{\textbf{Multi-label analysis of Prima}}
    \label{exfig:ex_data8}
\end{figure*}

\begin{figure*}[p!]
    \contcaption{\textbf{Multi-label analysis of Prima}
    \textbf{a}, We performed a multi-label classification analysis across all 52 diagnoses. The Prima logit-label matrix above shows the AUC value for each logit-label binary comparison. The diagnoses were ordered using consensus clustering for easier visualization \cite{Monti2003-bo}. The lower left-hand matrix is the correlation matrix of the ground truth labels with the same ordering. Prima learned the label co-occurrence relationships, such as ventriculomegaly often co-occurs with ventricular catheters or brain tumors co-occur with mass effect. It also correctly captures the semantic similarity of diagnoses within differential diagnoses. We observe higher AUC values for related structural pediatric diagnoses, brain tumor diagnoses, and brain injury/trauma. These findings demonstrate that Prima has correctly modeled the multi-label classification problem while learning the semantic relationships between related diagnoses. \textbf{b}, Multi-label LIME analysis shows that Prima attends to different pathologic regions of the same MRI sequence depending on the diagnostic prediction. High LIME scores are assigned to tokens within the posterior fossa when the LIME scores are computed for `medulloblastoma' prediction. Conversely, high LIME scores are assigned to the enlarged ventricles when the LIME scores are computed for 'ventriculomegaly' prediction. We see similar patterns investigating the relationship between `brain contusion' and `midline shift' labels, and `brain metastasis' and `vasogenic edema' labels. Prima demonstrates \emph{trusthworthy multi-label classification}.
    }
\end{figure*}
\clearpage

\begin{figure*}[p!]
    \centering\includegraphics[scale=0.77]{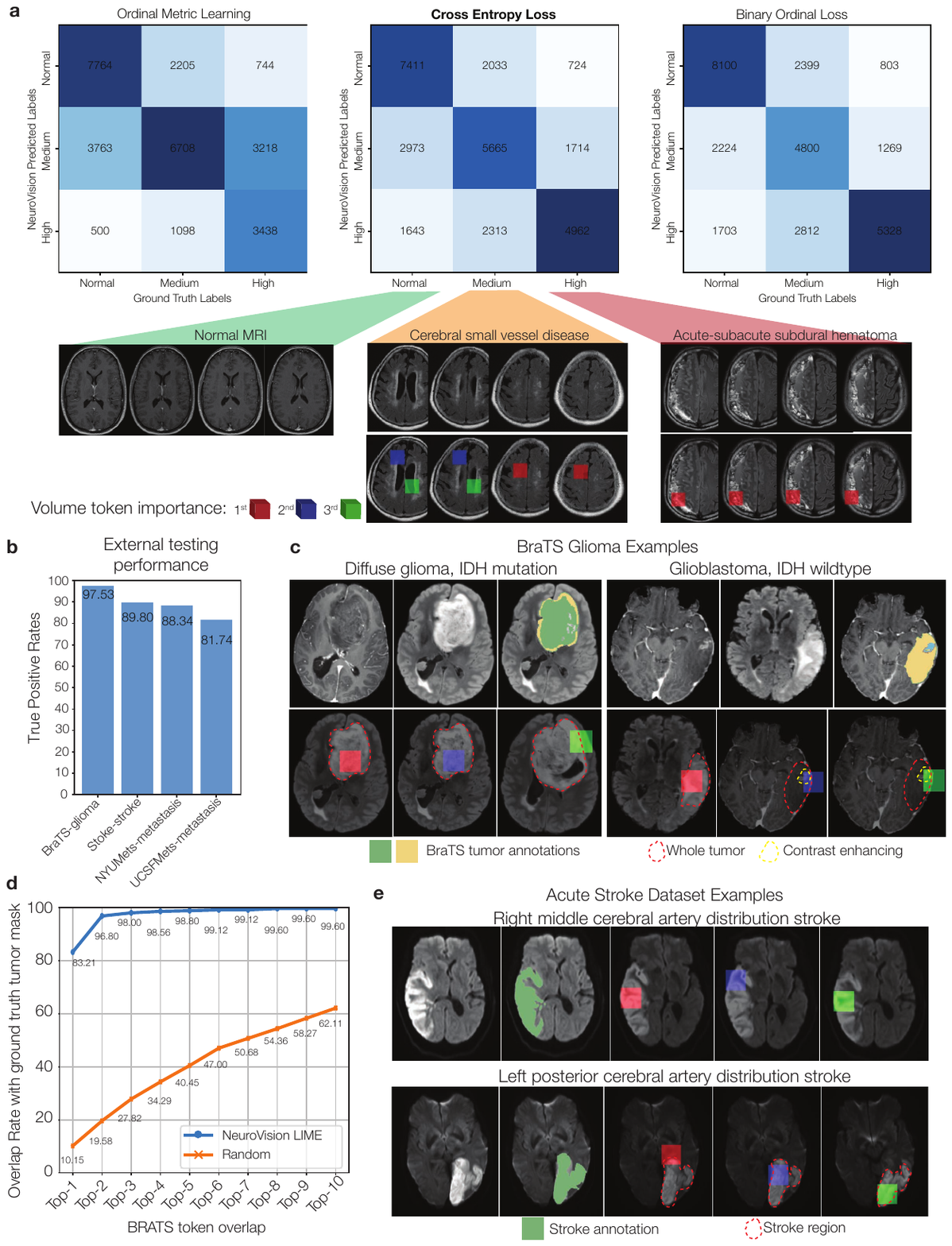}
    \caption{\textbf{Extended clinical task and transfer learning results}. Caption on next page.}
    \label{exfig:ex_data9}
\end{figure*}

\begin{figure*}[p!]
    \contcaption{\textbf{Extended clinical task and transfer learning results}
    \textbf{a}, Confusion matrices for ordinal metric learning, cross entropy, and binary ordinal losses for MRI prioritization. Cross entropy provided the overall best results. Examples of normal, medium, and high acuity are shown. Most importantly for triage and acuity assessment, misclassification rates were lowest for normal-high discrimination. \textbf{c}, External testing performance on BraTS, Stroke, NYUMets, and UCSFMets datasets. We see true positive rates on par with our prospective testing cohort. \textbf{c}, LIME importance scores extended to the BraTS dataset, identifying volume tokens within the externally annotated tumor regions. An example of both a lower grade diffuse glioma, IDH mutation, and a higher grade glioblastoma, IDH wildtype, MRI are shown. Prima correctly identified the FLAIR hyperintense regions as evidence of tumor infiltration. Contrast enhancing regions of the glioblastoma were most important for high grade glioma classification. \textbf{d}, Quantitative evaluation of LIME volume token selection versus the segmented tumor regions. We show that Top-3 accuracy for Prima selecting volume tokens within the ground truth segmented tumor region was 98.0\%. These results provide external, quantitative evaluation of trustworthy predictions from Prima. \textbf{b}, LIME visualizations for our external acute stroke dataset. An example of an acute middle cerebral artery distribution stroke and a posterior cerebral artery distribution stroke are shown. Prima assigned high LIME score to regions of diffuse restriction on DWI images, indicative of acute ischemia.
    }
\end{figure*}
\clearpage

\begin{figure*}[p!]
    \centering\includegraphics[scale=0.57]{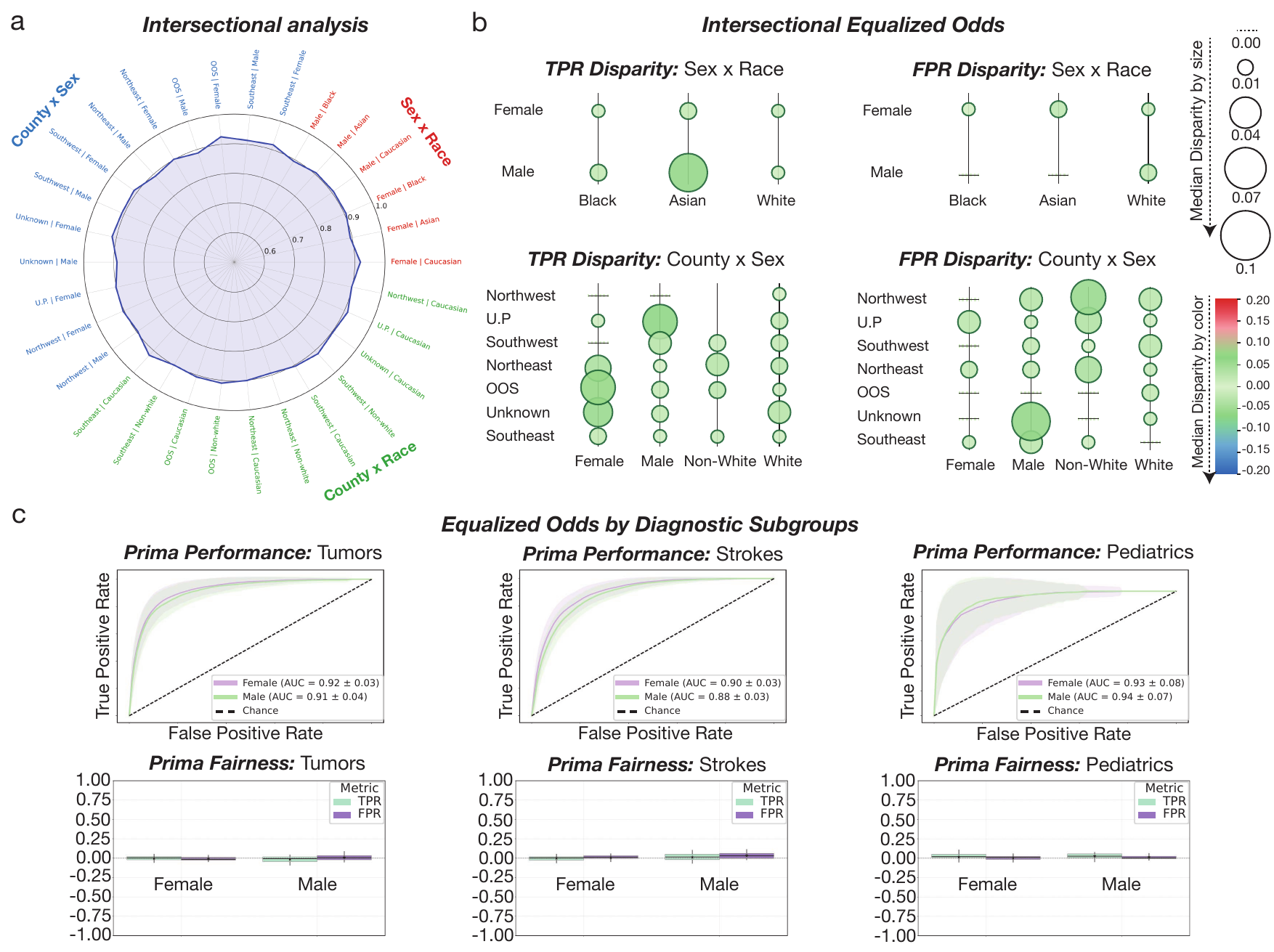}
    \caption{\textbf{Intersectional and equalized odds analysis}.}
    \label{exfig:ex_data10}
\end{figure*}

\begin{figure*}[p!]
    \contcaption{\textbf{Intersectional and equalized odds fairness analysis}
    \textbf{a}, Intersectional fairness analysis of sex, race, and geographic region on our prospective testing cohort that includes a diverse patient population. Radar plot shows each intersectional group and mean AUC values for each group. Prima shows minimal variance in diagnostic performance across intersectional groups. \textbf{b}, Intersectional equalized odds analysis demonstrate minimal disparity in TPR or FPR across intersectional groups. TPR and FPR disparity was less than 0.1 threshold we used for clinical significance. \textbf{c}, Equalized odds by diagnostic subgroup are shown, including brain tumors, strokes, and pediatric diagnoses. Similar to the demographic features shown in Fig. 2, there is minimal disparity between diagnostic subgroups. Prima displays algorithmic fairness on intersectional and equalized odds analysis of our diverse prospective testing cohort.
    }
\end{figure*}

\clearpage



\begin{figure*}
\vspace{-20pt}
\centering
\small
\begin{minipage}{\dimexpr\linewidth-2\fboxsep-2\fboxrule}
\begin{algorithm}[H]
\caption{Supplementary Data Figure 4: Random Token Permutation Strategy in PyTorch style}
\label{alg:clip}
\definecolor{codeblue}{rgb}{0.0,0.5,0.0}
\definecolor{codekw}{rgb}{0.85, 0.18, 0.50}
\lstset{
  basicstyle=\fontsize{10pt}{9pt}\selectfont,
  columns=fullflexible,
  breaklines=true,
  captionpos=b,
  commentstyle=\color{codeblue},
  keywordstyle=\color{codekw},
}
\begin{lstlisting}[language=python]
# vol_tokenizer: MRI tokenizer function, outputs N dimensional array of F dimensional 3D tokens
# p: function to permute tokens to desired shape
# mb: image minibatches
# V: VQVAE model
# S: set of all image minibatches

def get_randomly_permuted_tokens(mb):
    # Buckets for the 3 unique token shapes (e.g. 4x32x32, 32x4x32, and 32x32x4)
    token_buckets = {(4,32,32): [ ] , (32,4,32): [ ], (32,32,4): [ ] }

    for i in mb:
        # Tokenize each image in the minibatch and find shape of each token
        token_list, token_shape = vol_tokenizer(i)
        # Pass token list to corresponding dictionary pair based on token shape
        token_buckets[token_shape].extend(token_list)
    
    for key in tokens_buckets:
        # Stack all tokens in a given key, value pair
        token_buckets[key] = torch.stack(token_buckets[key])
    
    # Randomly select desired token shape for the minibatch
    selected_shape = random.choice(list(token_buckets.keys()))
    
    return token_buckets[selected_shape].permute(0,*random.shuffle([1,2,3]))

# Get dataset of all tokenized, permuted data
dt = Dataset(get_randomly_permuted_tokens(mb) for mb in Dataset(S))
dl = Dataloader(dt)

# Feed forward each minibatches through VQVAE model
for tokens in dl:
    loss = V(tokens)
    loss.backward()
    optimizer.step()

\end{lstlisting}

\end{algorithm}
\end{minipage}
\vspace{-10pt}
\end{figure*}

\begin{figure*}
\vspace{-20pt}
\centering
\small
\begin{minipage}{\dimexpr\linewidth-2\fboxsep-2\fboxrule}
\begin{algorithm}[H]
\caption{Supplementary Data Figure 5: MRI-Report Multimodal Representation Training Objective in PyTorch style}
\label{alg:clip}
\definecolor{codeblue}{rgb}{0.0,0.5,0.0}
\definecolor{codekw}{rgb}{0.85, 0.18, 0.50}
\lstset{
  basicstyle=\fontsize{10pt}{9pt}\selectfont,
  columns=fullflexible,
  breaklines=true,
  captionpos=b,
  commentstyle=\color{codeblue},
  keywordstyle=\color{codekw},
}
\begin{lstlisting}[language=python]
# f: volume tokenizer (3D-CNN)
# F: pixel intensity filtering function
# ViT_seq: sequence model (transformer)
# ViT_st: study model (transformer)
# G: report encoder (gpt-2 transformer, pre-trained on reports by )
# E_sn: sequence name encoder (transformer)
# E_stn: study name encoder (LSTM)
# S: minibatch of N MRI Studies
# R: minibatch of matched N MRI Reports
# P_patdis: the patient discrimination projection layer
# lambda_patdis: patient discrimination loss weight
# tau, tau_p: temperature for clip and patient discrimination loss

def encode_sequence(sequence):
   # Encode the sequence name from the sequence metadata
   seq_name_encoded = E_sn(sequence.name) 
   # Tokenize series, concatenate name, forward pass f
   return  ViT_seq(torch.cat([f(F(sequence)),seq_name_encoded])) 

def encode_study(study):
   # Encode the study name from the study metadata
   study_name_encoded = E_st(study.name) 
   # Encode sequences, concatenate name, forward pass g
   encoded_seqs = torch.stack([encode_sequence(sequence) for sequence in study])
   return  ViT_st(torch.cat([encoded_seqs,study_name_encoded])) 

def patient_discrimination_loss(studies):
   # compute projected sequence embedding for patient discrimination loss
   seq_embs = []
   seq_maps = [] # maps each seq emb to the study it belongs to
   for i,study in enumerate(studies):
      seq_embs.extend([P_patdis(encode_sequence(sequence)) for sequence in study])
      seq_map.extend([i for sequence in study])
   seq_embs = torch.stack(seq_embs)
   # compute positive mask from seq_map
   seq_maps = torch.LongTensor(seq_maps)
   mask = (seq_maps[:,None] == seq_maps)
   masksum = mask.sum(dim=1)
   logits = torch.matmul(seq_embs,seq_embs.t())/tau_p
   logits.fill_diagonal_(-10)
   q = F.softmax(logits,dim=1)
   aggscores = (mask * q).sum(dim=1)
   return -torch.dot(aggscores.log(),1.0/masksum)
   
# Compute the MRI study and report features 
S_features = torch.stack([encode_study(study) for study in S])
R_features = G(R)
# Compute inner product between all MRI and Report features
logits = torch.matmul(S_features, R_features.t())*exp(tau)
# Targets: diagonal elements are the matching pairs
targets = torch.arange(logits.size(0))
# Cross-entropy loss for both mri-to-report and report-to-mri
loss_mri2report = F.cross_entropy(logits, targets)
loss_report2mri = F.cross_entropy(logits.t(), targets)
# Total loss is the average of both losses
loss_clip = (loss_mri2report + loss_report2mri) / 2
loss_patdis = patient_discrimination_loss(S)
loss = loss_clip + lambda_patdis * loss_patdis
\end{lstlisting}
\end{algorithm}
\end{minipage}
\vspace{-10pt}
\end{figure*}

\renewcommand{\figurename}{Supplementary Data Figure}
\setcounter{figure}{5}

\begin{figure*}[p!]
    \centering\includegraphics[scale=0.75]{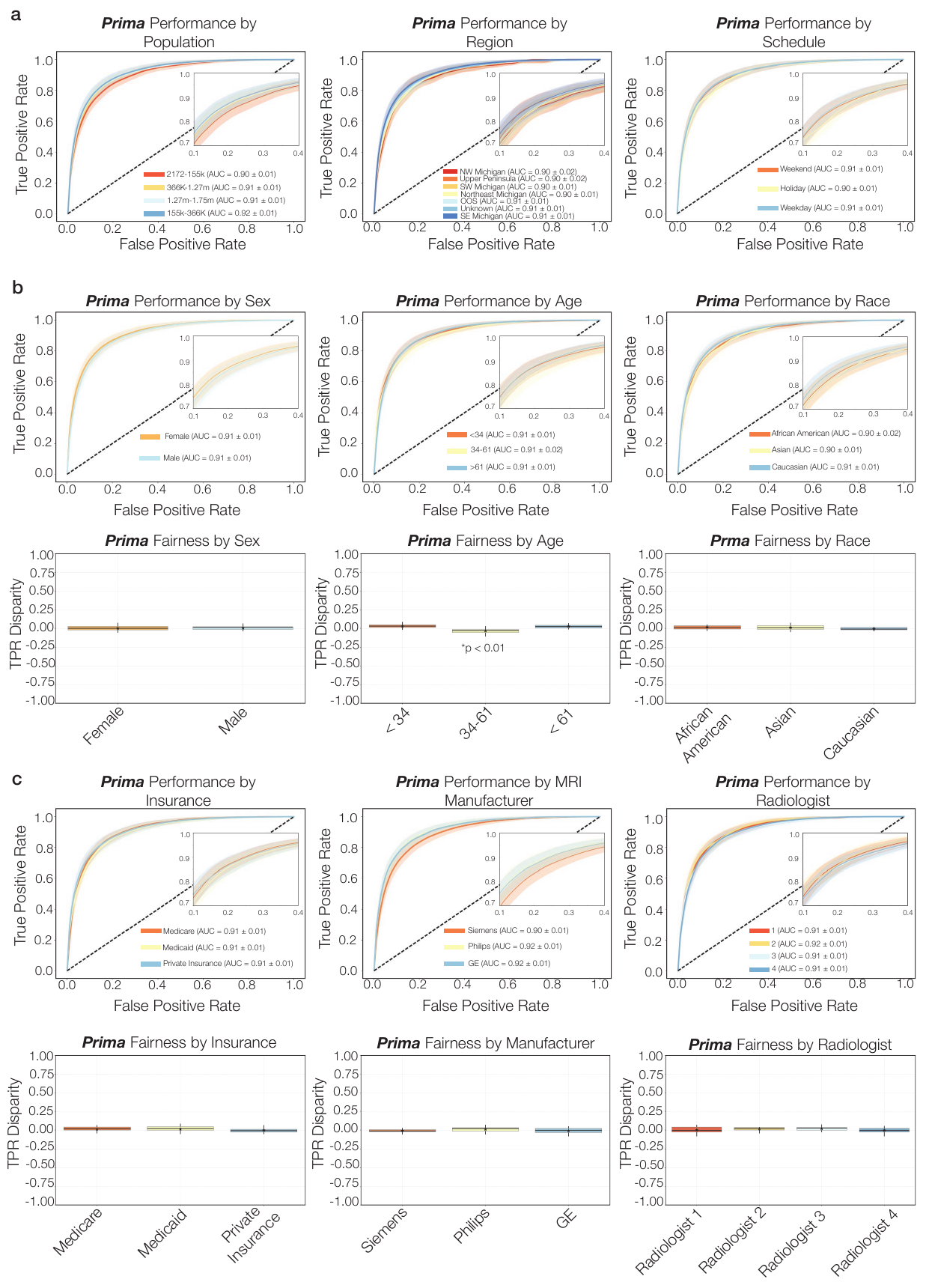}
    \caption{\textbf{Subgroup performance and fairness analysis}. Caption on next page.}
    \label{supfig:sup_data6}
\end{figure*}

\begin{figure*}[p!]
    \contcaption{\textbf{Subgroup performance and fairness analysis} \textbf{a}, Prima performance for tasks presented in Fig. 4: population, region, and schedule. \textbf{b}, Prima performance and fairness analysis for non-modifiable sensitive attributes: sex, age, and race. Prima showed consistently high performance, on par with the population performance, across all subgroups. Prima demonstrated algorithmic fairness for sex and race; we found a statistically significant difference in TPR disparity for ages between 34-61. These may be related to greater diversity of diagnoses within this subgroup compared to pediatric and older populations. \textbf{c}, Prima performance and fairness analysis for modifiable sensitive attribute: insurance status, MRI manufacturer, and radiologist. Prima did not show statistically significant difference in performance or fairness based on government-sponsored versus private insurers. Importantly, MRI performance was consistent across the three major MRI manufacturers: Seimens, Philips, and GE. P-values are greater than 0.05 with multiple hypothesis correction unless otherwise noted.}
\end{figure*}

\clearpage
\begin{figure*}[p!]
    \centering\includegraphics[scale=0.75]{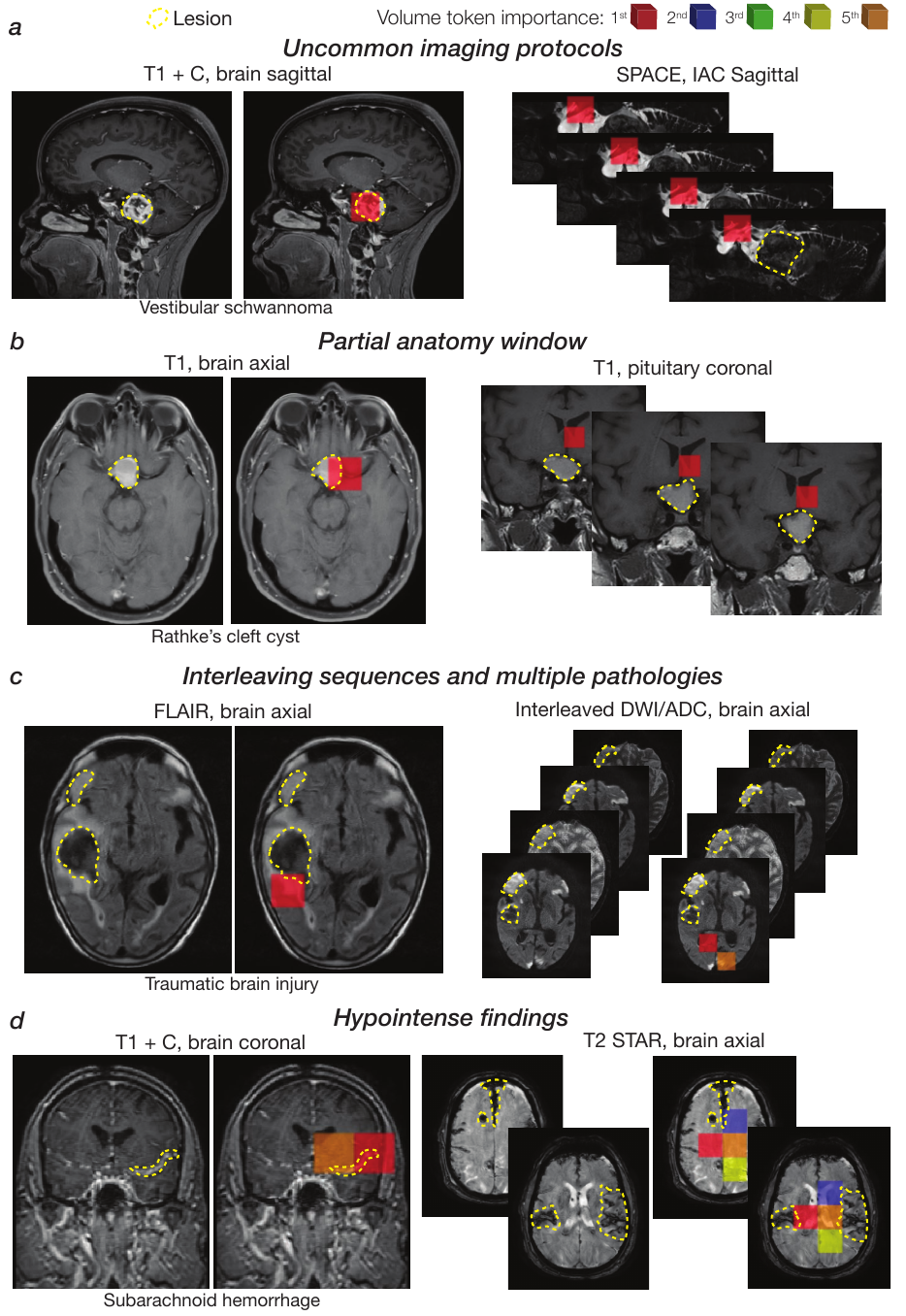}
    \caption{\textbf{Error analysis with LIME visualizations}. Caption on next page.}
    \label{supfig:sup_data7}
\end{figure*}

\begin{figure*}[p!]
    \contcaption{\textbf{Error analysis with LIME visualizations} We present several illustrative cases that demonstrate some of the limitations of Prima. \textbf{a}, A vestibular schwannoma in the cerebellopontine angle in shown and recognized by Prima on a contrasted T1-weighted sagittal brain image. However, a SPACE IAC protocol shows that Prima is unable to identify the lesion. This is due to SPACE protocols being relatively uncommon in UM-220K and poor image contrast between the brain parenchyma and the lesion. \textbf{b}, A Rathke's cleft cyst is identified by Prima on the T1-weighted brain axial image. We have found that Prima can be limited by imaging protocols that include partial views of the brain, similar to IAC protocol. We believe that this is due to limitations of standard vision transformers positional embeddings, which do not encode anatomic information. This is an active area of research we are pursuing. \textbf{c}, A patient with a severe traumatic brain injury is shown and a brain MRI was acquired for prognostication. The patient has several intracranial hemorrhages and regions of ischemia. Interleaved DWI and ADC sequences can result in poor localization of ischemic lesions by Prima, especially in the context of multiple pathologies. \textbf{d}, A patient with a subarachnoid hemorrhage is shown. Prima correctly identified subarachnoid blood in the left sylvian fissure on a contrasted T1-weighted image. However, Prima stuggles to identify acute blood on T2* sequence due to hypointensity of the blood products. Detecting dark objects is a recognized problem in computer vision.
    } 
\end{figure*}

\end{document}